\documentclass[letterpaper]{article} 
\usepackage{aaai2026}  
\usepackage{times}  
\usepackage{helvet}  
\usepackage{courier}  
\usepackage[hyphens]{url}  
\usepackage{graphicx} 
\usepackage{natbib}  
\usepackage{caption} 
\usepackage{algorithm}
\usepackage{algorithmic}
\usepackage{multirow}

\usepackage{newfloat}
\usepackage{listings}
\DeclareCaptionStyle{ruled}{labelfont=normalfont,labelsep=colon,strut=off} 
\floatstyle{ruled}
\newfloat{listing}{tb}{lst}{}
\floatname{listing}{Listing}
\title{Marmot: Object-Level Self-Correction via Multi-Agent Reasoning}

\author{%
  \textbf{Jiayang Sun}\textsuperscript{\rm 1, 2},
  \textbf{Hongbo Wang}\textsuperscript{\rm 1, 2},
  \textbf{Jie Cao}\textsuperscript{\rm 1, 2},
  \textbf{Huaibo Huang}\textsuperscript{\rm 1, 2},
  \textbf{Ran He}\textsuperscript{\rm 1, 2}
}

\affiliations{
    \textsuperscript{\rm 1}MAIS \& NLPR, Institute of Automation, Chinese Academy of Sciences, Beijing, China\\
    \textsuperscript{\rm 2}School of Artificial Intelligence, University of Chinese Academy of Sciences, Beijing, China
}

\usepackage{bibentry}

\begin{document}


\twocolumn[
{%
\renewcommand\twocolumn[1][]{#1}
\maketitle
\begin{center}
\centering
\includegraphics[width=0.95\textwidth]{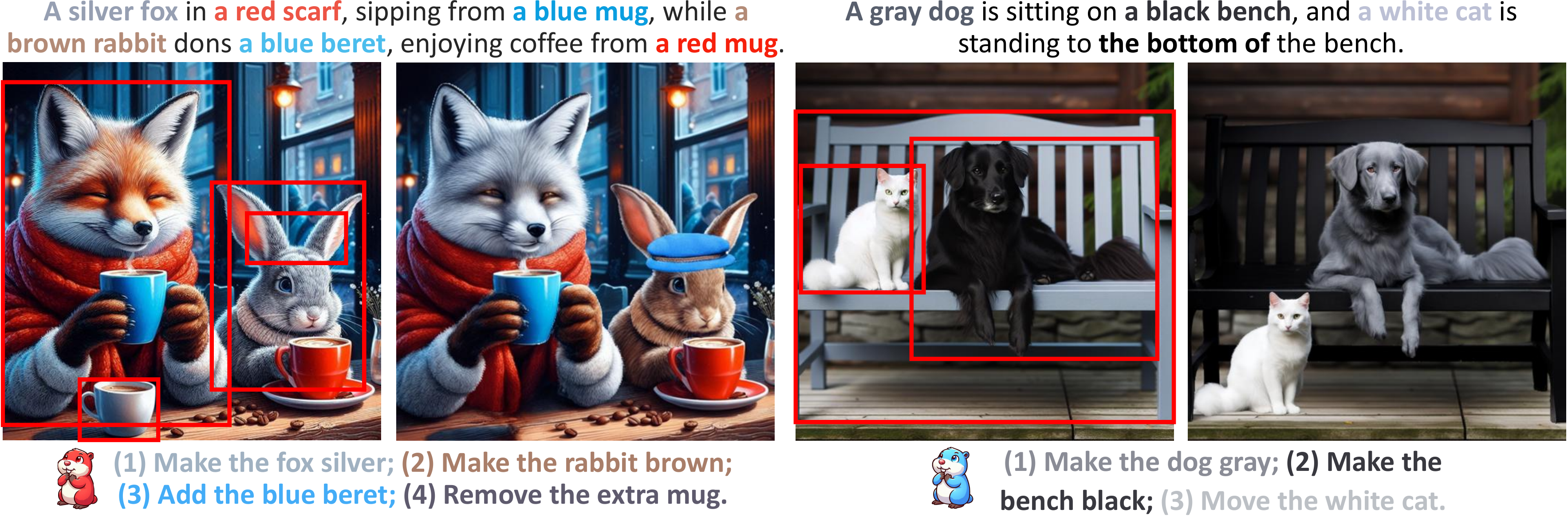}
{\captionsetup{hypcap=false}  
\captionof{figure}{Existing diffusion-based models often struggle with image-text inconsistencies in object counting, attributes, and spatial relationships. Our \textbf{Marmot} can seamlessly integrate with text-to-image models to facilitate object-level self-correction, ensuring precise alignment with user descriptions in object counting, attributes, and spatial relationships.}
\label{figure:teaser}}
\end{center}
}]

\begin{abstract}

While diffusion models excel at generating high-quality images, they often struggle with accurate counting, attributes, and spatial relationships in complex multi-object scenes.
One potential solution involves employing Multimodal Large Language Model (MLLM) as an AI agent to construct a self-correction framework. However, these approaches heavily rely on the capabilities of the MLLMs used, often fail to account for all objects within the image, and suffer from cumulative distortions during multi-round editing processes.
To address these challenges, we propose \textbf{Marmot}, a novel and generalizable framework that leverages \textbf{M}ulti-\textbf{A}gent \textbf{R}easoning for \textbf{M}ulti-\textbf{O}bjec\textbf{t} Self-Correcting to enhance image-text alignment.
First, we employ a large language model as an Object-Aware Agent to perform object-level divide-and-conquer, automatically decomposing self-correction tasks into object-centric subtasks based on image descriptions.
For each subtask, we construct an Object Correction System featuring a decision-execution-verification mechanism that operates exclusively on a single object's segmentation mask or the bounding boxes of object pairs, effectively mitigating inter-object interference and enhancing editing reliability.
To efficiently integrate correction results from subtasks while avoiding cumulative distortions from multi-stage editing, we propose a Pixel-Domain Stitching Smoother, which employs mask-guided two-stage latent space optimization. This innovation enables parallel processing of subtasks, significantly improving runtime efficiency while preventing distortion accumulation.
Extensive experiments demonstrate that Marmot significantly improves accuracy in object counting, attribute assignment, and spatial relationships for image generation tasks.

\end{abstract}

\section{Introduction}
\label{sec:intro}

Diffusion-based generative models \cite{dalle1,dalle2,dalle3,imagen,ldm,sdxl} have demonstrated exceptional capabilities in producing high-quality images. However, when tasked with generating complex scenes that contain multiple distinct objects, these models often struggle with counting, attribute binding, and spatial relationships between different objects. This limitation leads to outputs that do not fully align with the user’s intended descriptions. As shown in Figure~\ref{figure:teaser}, on the left, the image generated by DALL-E 3 \cite{dalle3} inaccurately depicts the colors of the fox and the rabbit, omits a beret, and generates a additional mug. The right image further demonstrates chaotic color assignments, resulting in a gray bench and black dog, along with an incorrect spatial relationship between the cat and the bench.


Recent studies \cite{boxdiff,controlnet,densediffusion,freecontrol,gligen,instancediffusion,migc,multidiffusion,attendandexcite,lmd,reco} have introduced mechanisms such as boundary layouts and localized text prompts through users or Large Language Models (LLMs), in order to better align generative models with user intent, thereby enhancing control over image composition. This paradigm effectively constructs scenes but faces challenges in preserving the unique attributes of individual objects. Furthermore, the hard constraint of generating specific objects at predefined locations often introduces noticeable artifacts, degrading both the realism and fine-grained details of the synthesized images.

Another potential line of research \cite{sld, rpg, genartist, divide} has proposed a framework that utilizes a Multimodal Large Language Model (MLLM) as an AI agent to integrate domain-specific models and iteratively refine images based on user intent. This approach employs a training-free post-processing self-correction framework and can be seamlessly integrated with robust text-to-image models like DALL-E 3 \cite{dalle3} or even innovative models that will be developed in the future. Therefore, this approach generates images with enhanced realism and detail. However, these frameworks face two critical limitations: (1) They rely on a single agent, heavily dependent on the MLLM’s capabilities, and often fail to simultaneously handle all objects, leading to partial corrections; (2) They neglect to account for the cumulative distortions introduced by their serial editing pipeline, which progressively degrades multi-object alignment.


\begin{figure}[t]
   \centering
   \includegraphics[width=0.47\textwidth]{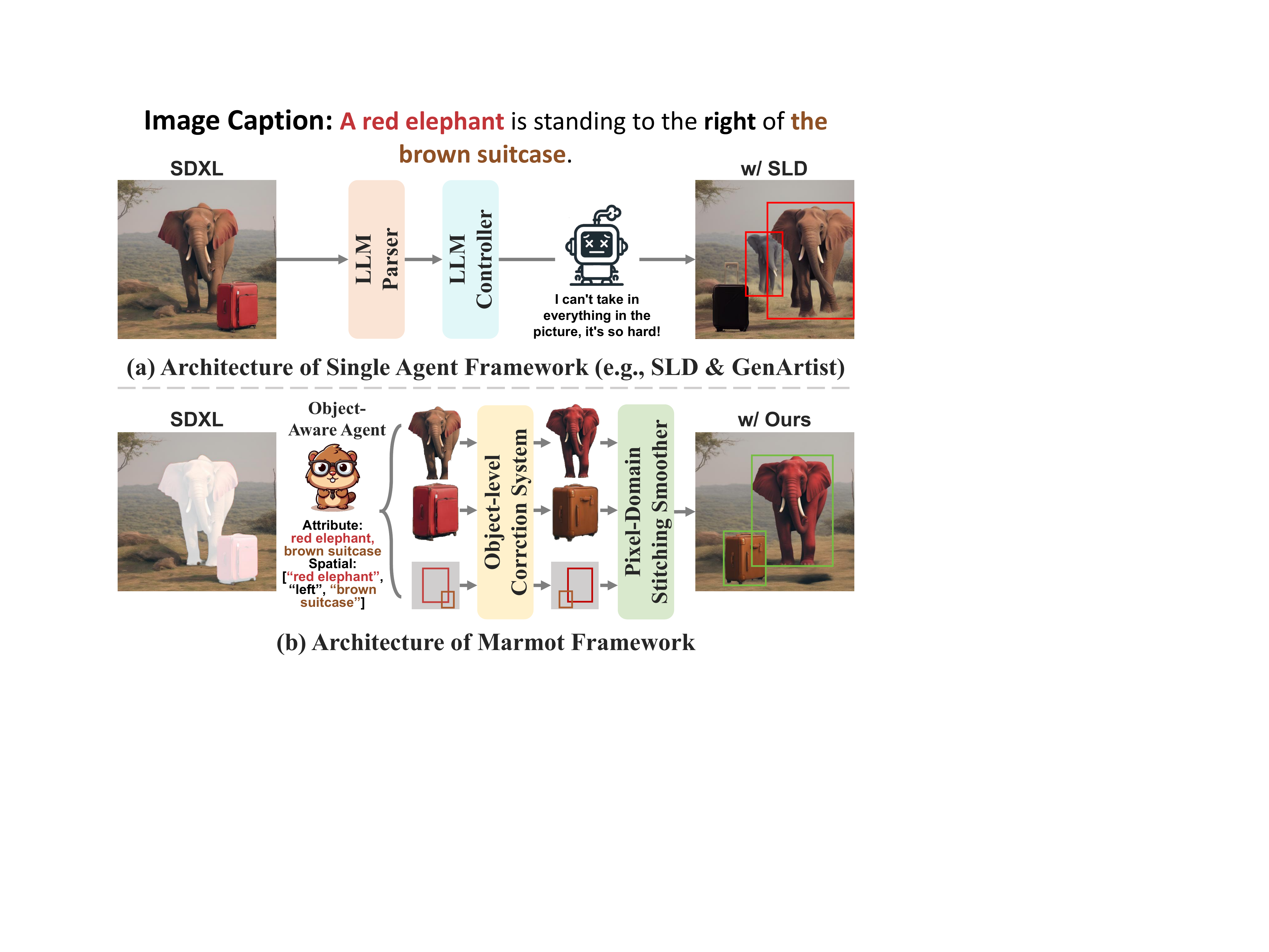} 
   \caption{\textbf{Comparison Between Single Agent Framework and Marmot.} By decomposing the self-correction into object-level subtasks, OAA enables OCS to process only a single object's segmentation mask or bounding boxes of object pairs at a time. This approach significantly relaxes the capability requirements for the MLLM.}
   \label{figure:compare}
\end{figure}

To address these challenges, we introduce Marmot, a versatile framework that leverages multi-agent reasoning to achieve object-level self-correcting.
Inspired by the ideas of ``divide and conquer'' \cite{migc} and ``problem decomposition'' \cite{cot,mmcot,tot}, we innovatively integrate chain-of-thought reasoning into image self-correcting tasks. 
Notably, the LLM-based \textit{Object-Aware Agent} (OAA), acting as the central planner, autonomously analyzes user intent to decompose the overall image correction task into a sequence of manageable, object-specific subtasks.
For each subtask, as shown in Figure~\ref{figure:compare}, the \textit{Object Correction System} (OCS) processes only the segmentation mask of a single object or the bounding boxes of object pairs. This approach fundamentally differs from whole-image processing frameworks like SLD \cite{sld}, effectively isolating the correction process and thus preventing inter-object interference.
Internally, OCS operates a dedicated \textit{decision-execution-verification} mechanism, consisting of three specialized components: Decision-Makers, Executors, and Verifiers. This mechanism significantly enhances the reliability of the editing process.

Our object-level automatic decomposition approach not only effectively reduces the reliance on agent capabilities, but also yields more reliable outcomes. However, two critical challenges arise: computational inefficiency and subtask result integration. To alleviate these issues, we propose \textit{Pixel-Domain Stitching Smoother} (PDSS). We first erase objects that need modification by Remove Anything \cite{inpaintanything} , and then paste the corrected results of each subtask into the erased image. Finally, we invert the synthesized image into the latent space and refine it with SDXL-Refiner \cite{sdxl} in a two-stage optimization: In the first \textit{K} iterations, latent features are multiplied by a composite mask to enhance details of the target objects while preserving the high-frequency detail information of the background.
Subsequently, \textit{T} -- \textit{K} mask-free global refinement iterations blend object boundaries into the background.
This approach enables parallel execution of subtasks, eliminating runtime bottlenecks while preventing image distortion caused by multi-stage editing processes.

Our comprehensive evaluation shows that Marmot successfully enhances the performance of text-to-image models on the T2I-CompBench \cite{t2icompbench}, especially in terms of attribute binding and object relationships. Specifically, SDXL \cite{sdxl} achieves improvements of 11.76\%, 6.86\%, and 5.47\% in color, shape, and texture tasks, respectively. In spatial relationship tasks, Marmot also delivers significant performance gains, boosting SDXL and DALL-E 3 \cite{dalle3} by 13.12\% and 8.08\%. These results highlight Marmot's versatility and effectiveness across multiple models and datasets.

Our contributions are summarized as follows:

\begin{itemize}

\item[1)] We propose an LLM-based OAA that autonomously plans self-correction tasks into object-centric subtasks. For each subtask, OCS processes only a single object's segmentation mask or an object pair, performing corrections through a decision-execution-verification process. This mechanism inherently eliminates inter-object interference while enhancing editing robustness. 

\item[2)] We present PDSS that seamlessly integrates subtask results through mask-guided two-stage latent space optimization. The module enhances computational efficiency by enabling parallel execution of subtasks, while effectively preserving high-frequency detail information in the image background and avoiding distortion artifacts that arise from multi-stage editing pipelines.

\item[3)] Our framework effectively addresses persistent challenges in complex multi-object scene generation. Extensive experiments demonstrate that Marmot significantly enhances image-text alignment and outperforms state-of-the-art self-correction frameworks.
\end{itemize}

\section{Related work}
\label{sec:Related work}

\subsection{Text-to-Image Generation}
The field of Text-to-Image (T2I) generation using diffusion models (DM) \cite{ddpm} has advanced rapidly. Models such as DALL-E \cite{dalle1,dalle2,dalle3} achieve high-quality image synthesis by integrating image and text features from CLIP \cite{clip} and leveraging large-scale training and cascaded diffusion. Similarly, Imagen \cite{imagen} utilizes T5 \cite{t5} to enhance its performance in this domain. To boost efficiency, Latent Diffusion Model (LDM) \cite{ldm} applies diffusion in the latent space of VAEs \cite{vae}, while SDXL \cite{sdxl} further improves upon LDM with an expanded U-Net \cite{unet}, dual text encoders, and enhanced control for high-resolution details. Most recently, FLUX \cite{flux} and SD 3 \cite{sd3} adopt the Diffusion Transformer \cite{dit} as the base structure and employ flow matching techniques for training, establishing new state-of-the-art results through improved stability and sample efficiency in large-scale diffusion models.
Despite these advancements, T2I models often struggle with attribute and spatial consistency. Our framework performs object-level self-correction on images through multi-agent collaborative planning and reasoning, enhancing the image's capabilities in counting, attribute matching, and spatial relationship alignment.

\subsection{Image generation and editing with LLMs}
The remarkable performance of LLMs elevates the field of image generation to unprecedented levels. 
\cite{layoutgpt,layoutllm-t2i,lmd,sld,rpg} utilize the inferential capabilities of LLMs for layout planning, enhancing the spatial relationships and counting ability of generated images. 
\cite{diffusiongpt,suradapter,idea2img,rpg} optimize text prompts through LLMs, guiding T2I models to generate rich and detailed images.  
\cite{sld,rpg,genartist,hallo3d,divide} are the most similar to our framework. They leverage LLMs and MLLMs as AI-driven agents to facilitate image self-correction processes, thereby producing visual outputs that exhibit enhanced alignment with user intentions. Notably, SLD \cite{sld} employs an LLM controller to guide diffusion models through iterative self-correction, utilizing latent space operations such as addition, deletion, and repositioning to rectify inaccuracies in the generated images. RPG \cite{rpg} leverages multi-round iterative optimization, using LLMs to assess and provide feedback on the generated images. GenArtist \cite{genartist} treats LLMs as an ``artist'', decomposing user instructions, planning generation and editing steps, and validating and self-correcting at each stage to achieve a unified functionality for both image generation and editing.

The major differences between our approach and existing methods are: (1) an LLM-based OAA automatically decomposes self-correction tasks into object-level subtasks, restricting OCS's input to single-object segmentation masks instead of full images; (2) a multi-agent framework with decision-execution-verification mechanisms to enhance accuracy and robustness; (3) a PDSS that ensures seamless integration of subtask outputs, enabling parallel execution while avoiding distortion artifacts.
\section{Method}
\label{sec:method}


\begin{figure*}[t]
	\centering
	\includegraphics[width=1\textwidth]{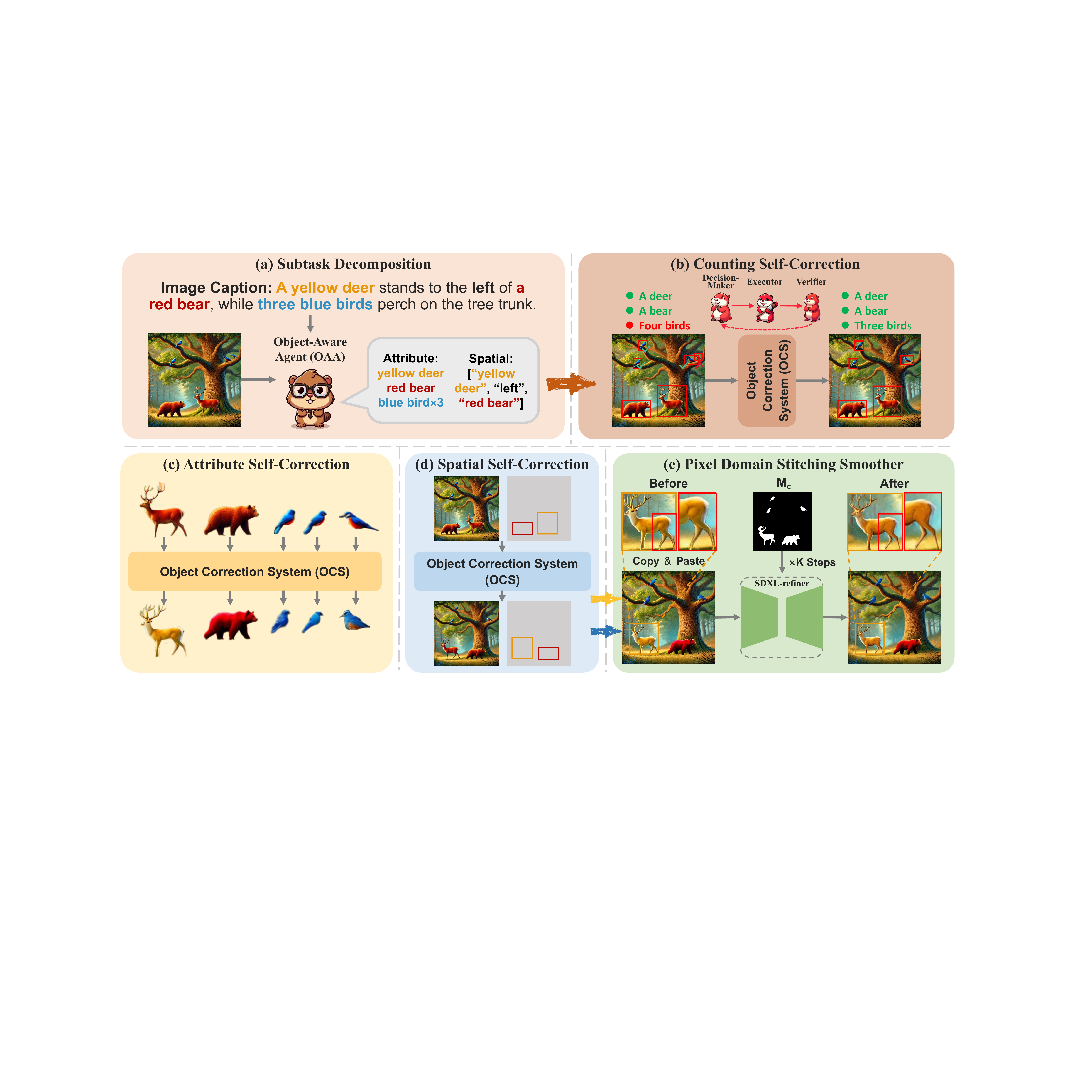} %
	\caption{\textbf{Overview of Marmot.} First, OAA  autonomously decomposes the self-correction task into object-level subtasks. Subsequently, OCS serially performs counting correction, while processing attribute and spatial corrections in parallel via decision-execution-verification mechanism. Finally, PDSS stitches the subtask results together to produce the refined image.}
	\label{figure:pipeline}
\end{figure*}

Our Marmot framework, as illustrated in Figure~\ref{figure:pipeline}, is designed to enhance image-text alignment, particularly in complex multi-object scenarios. Initially, Object-Aware Agent (OAA) interprets user intent and decomposes the task into object-level subtasks. Subsequently, Object Correction System (OCS) processes segmentation masks for individual objects or bounding boxes for object pairs with the decision-executor-verification mechanism. Finally, we introduce Pixel-Domain Stitching Smoother (PDSS) to seamlessly integrate the results of subtasks, improving efficiency and preventing distortions that arise from multi-stage editing. 
Furthermore, a comprehensive flowchart and the corresponding prompts and responses of the agents for each stage are included in the supplementary materials.
The implementation details of each module are presented below.

\subsection{Object-Aware Agent}
\label{OAA}
Our object-level self-correction workflow begins by analyzing the user intent to decompose the image self-correction task into object-level subtasks. We leverage an LLM-base OAA to extract three critical elements from user description: the number of target objects, their attributes, and spatial relationships between the specified objects. This enables category-based decomposition of the self-correction task. To ensure clarity and avoid ambiguity caused by repeated object names in subsequent processing, we assign distinct numerical suffixes to each object instance. As shown in Figure~\ref{figure:pipeline}, we distinguish identical objects using this scheme (e.g., \texttt{yellow dear\_1, red bear\_2, blue bird\_3, blue bird\_4, blue bird\_5}). Guided by a meticulously engineered prompt, the OAA outputs the extracted information in a structured format: object attributes and counting are represented as \texttt{(["Attribute Object\_ID"])}, while spatial relationships between objects are represented as \texttt{(["Object\_ID", "Spatial", "Object\_ID"])}. Here, ID denotes the numerical suffix identifying a specific object instance.

\subsection{Object Correction System}
\label{system}

Employing a single agent to determine attributes and spatial relations in images faces inherent limitations in attending to all object information. In our system, we address this by focusing on a single object’s segmentation mask or the bounding boxes of object pairs, which significantly reduces the complexity of self-correction tasks. Our multi-agent framework comprises three core roles: \textit{Decision-Maker}, \textit{Executor}, and \textit{Verifier}. These specialized agents are implemented by distinct foundation models for counting, attribute, and spatial self-correction and are detailed in subsequent sections.

\begin{figure}[t]
   \centering
   \includegraphics[width=0.47\textwidth]{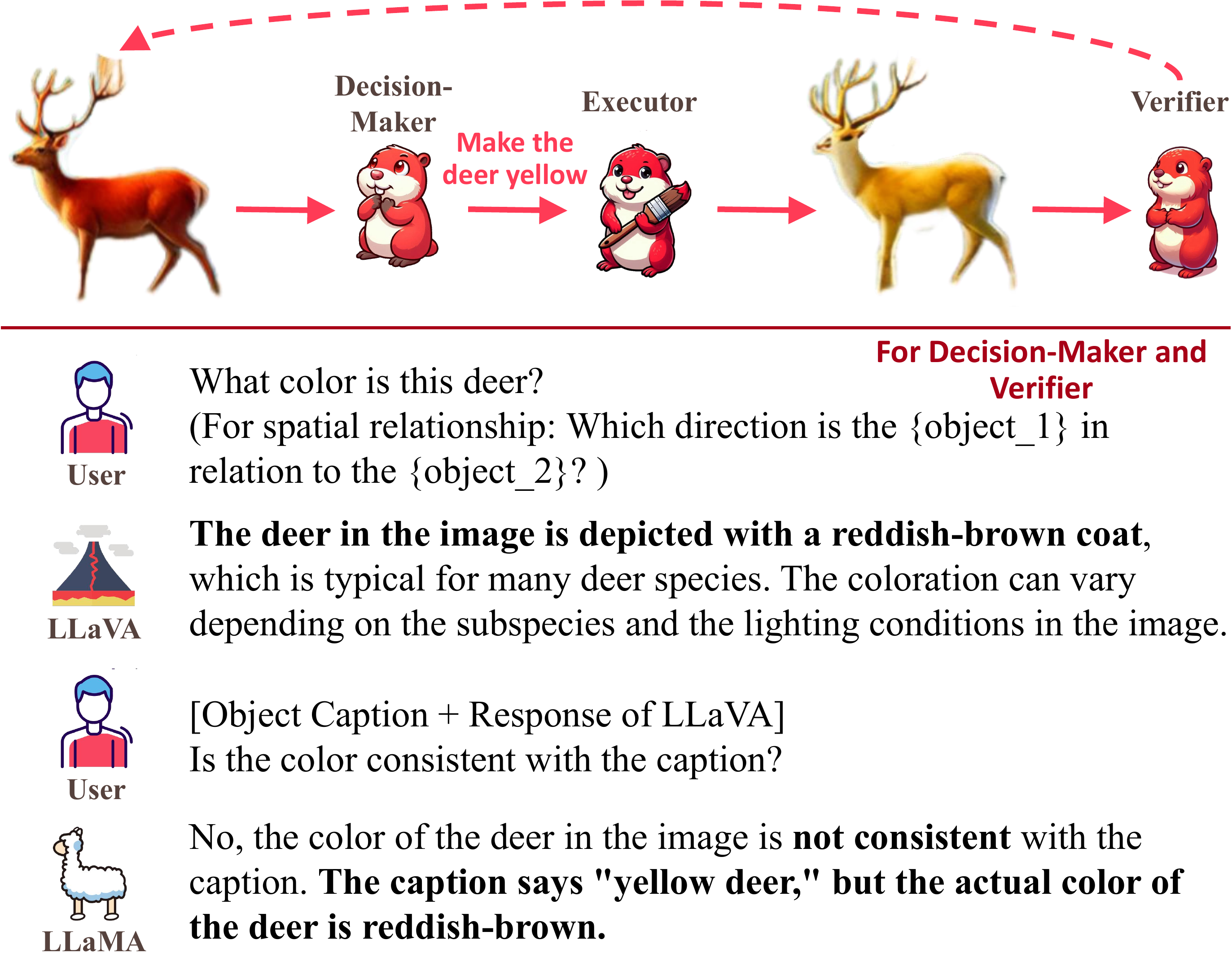}
   \caption{\textbf{OCS case}: decision-execution-verification mechanism for attribute self-correction of ``yellow deer".}
   \label{figure:system}
\end{figure}

\textbf{Decision-Maker} addresses the critical challenge of detecting attribute and spatial relationship errors in our multi-agent self-correcting system. For counting self-correction, we sort the bounding boxes based on the scores provided by Grounding-DINO \cite{groundingdino} to determine which bounding box is redundant. Excess boxes indicate overproduced objects, while missing boxes
suggest omissions. And for attribute and spatial self-correction, we adopt a two-stage visual reasoning paradigm to enhance error detection accuracy drawing on Multimodal Chain-of-Thought (MMCoT) reasoning approaches \cite{mmcot,ccot,ddcot}.
As shown in Figure~\ref{figure:system}, in the first stage, we leverage an MLLM to inquire about the object's attributes. In the second stage, we synthesize the target attribute with the MLLM response, feeding them to an LLM for comprehensive semantic alignment analysis. If the LLM responds with \texttt{No}, the Decision-Maker will generate the corresponding editing instruction: \texttt{Make the \{object\} \{attribute\}.}
Notably, spatial self-correction tasks require only adjustments to the first stage's query design, demonstrating our framework's task adaptability. In contrast to conventional multimodal understanding task, we establish a distinctive Image-Text-Image paradigm with the ultimate goal of generating high-quality editing instructions to drive the final image correction.

\begin{figure*}[t]
	\centering
	\includegraphics[width=1\textwidth]{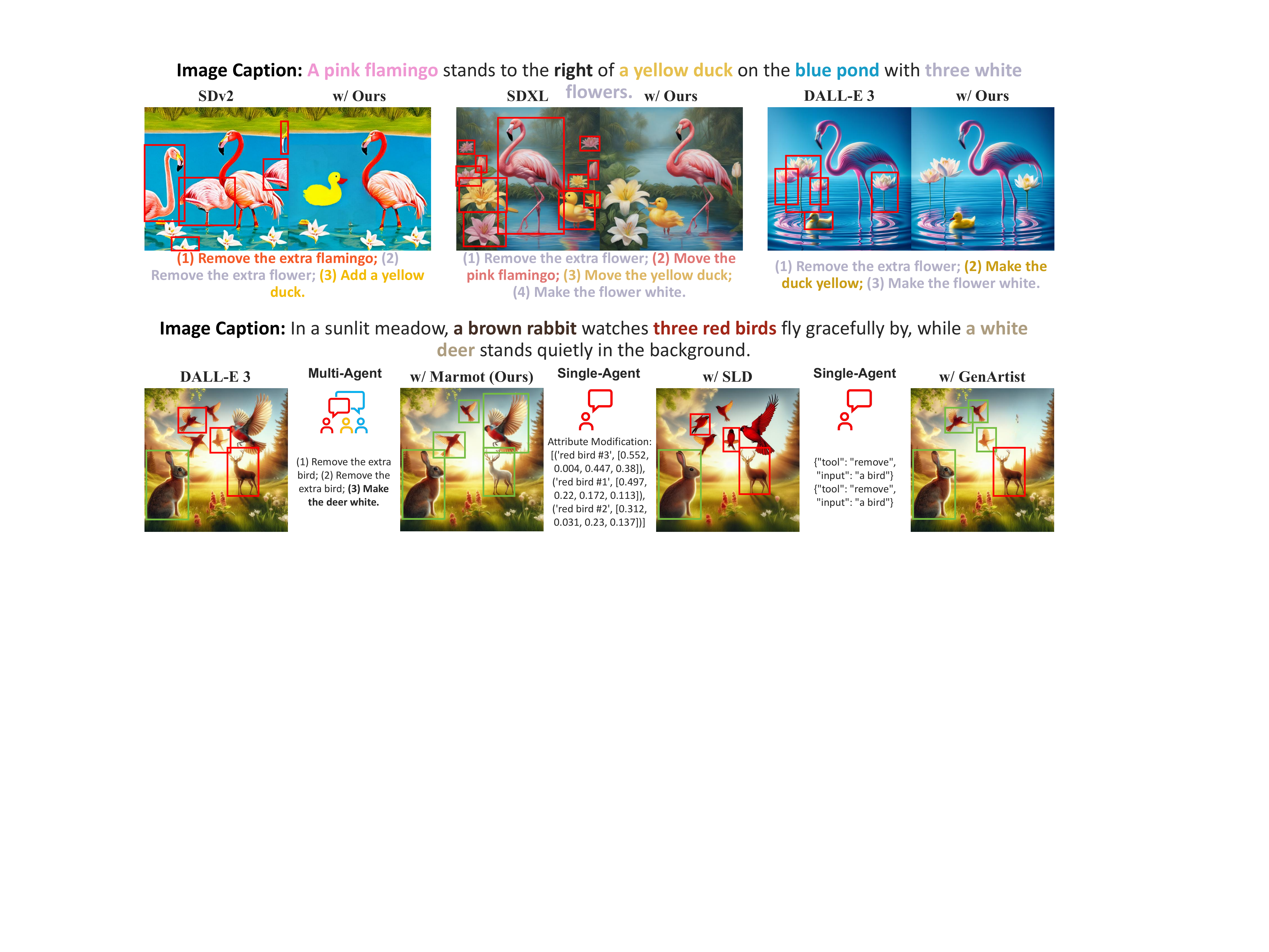} %
	\caption{\textbf{Visual Comparisons.} Incorporating our framework enables precise modeling of object counting, attributes, and spatial relationships, ensuring images comprehensively align with prompts. Through our OAA and OCS, Marmot achieves balanced attention across all objects, making it uniquely effective for complex cases like the ``white deer".}
	\label{figure:experiments}
\end{figure*}

\textbf{Executor} serves as the core component for executing image editing strategies formulated by the Decision-Maker. Task-specific diffusion models are conceptualized as functional agents within our framework. For counting self-correction, we follow Inpainting Anything \cite{inpaintanything}, using SAM \cite{sam} and LAMA \cite{lama} to remove redundant objects. To reintroduce missing objects, we provide their descriptions and existing bounding boxes to an LLM to attain new bounding box, and then generate the objects at specified locations by leveraging GLIGEN \cite{gligen} and AnyDoor \cite{anydoor}. For attribute self-correction, we leverage UltraEdit \cite{ultraedit} for image editing. These agents comprehend the editing strategies to optimize the images, ultimately producing refined visual objects. For spatial self-correction, the Executor employs an LLM that leverages bounding box annotations and spatial relationship descriptions to rationally adjust the positional configurations between target subjects. 

\textbf{Verifier} evaluates whether edited objects meet user requirements. When mismatches are detected, it triggers object re-correction. Since the input patterns and core tasks are identical between the Decision-Maker and Verifier, we implement the same object detective model and two-stage visual reasoning paradigm for verification. To prevent infinite cycles in the decision-execution-verification mechanism, the system enforces termination through a maximum iteration threshold. Notably, our decision-execution-verification pipeline operates with explicit transparency and controllability. Incorporating human evaluation and feedback further improves image quality\textemdash implementation details are in the supplementary material.

\subsection{Pixel Domain Stitching Smoother}

We propose PDSS to tackle the prevalent issues of background blurring and distortion in multi-stage editing, along with efficiency bottlenecks caused by serial editing processes. Marmot first employs LAMA \cite{lama} to erase target regions in the base image, followed by direct stitching of subtask editing results into the erased image. Subsequently, we discover that SDXL-Refiner \cite{sdxl} effectively smooths target edges and enhances object details during the refinement process. However, this model exhibits a measurable degree of interference with background regions during implementation. To resolve this issue, we innovatively invert synthesized images into latent space and implement a composite mask-guided two-stage optimization with SDXL-Refiner: During the first \textit{K} iterations, latent feature updates are constrained by composite mask ${\bf{M}}_\mathsf{c}$ to enhance structural details of while preserving the high-frequency information of the background. In this case, the latent is represented by $\hat{\mathbf{Z}}_{t}$, which is guided by the source latent ${\mathbf{Z}}_{t}$ and composite mask ${\bf{M}}_\mathsf{c}$:
\[
   \hat{\mathbf{Z}}_{t} = 
   \hat{\mathbf{Z}}_{t}
   \odot 
   {\bf{M}}_\mathsf{c} 
   + 
   {\bf{Z}}_{t}
   \odot 
   (1-{\bf{M}}_\mathsf{c}).
\]
Subsequent \textit{T} -- \textit{K} iterations perform global optimization without ${\bf{M}}_\mathsf{c}$ to ensure seamless integration of the object into the background. PDSS breaks through efficiency limitations via parallel processing, achieving edge smoothing and detail enhancement while maintaining background integrity.
\section{Experiments}
\label{sec:experiments}

\renewcommand{\arraystretch}{1.2}
\setlength{\tabcolsep}{1mm} 
\begin{table*}[]
\caption{\textbf{Quantitative Comparison on T2I-CompBench.} Our method demonstrates exceptional self-correction capabilities in attribute binding and object relationships. Notably, by employing AI agents with only 8B-parameter for image planning, we achieve performance that surpasses the GPT-4-based self-correcting framework.}
\label{table:evaluation}
\centering
\begin{tabular}{ccccccc}
\hline
\multicolumn{1}{c}{\multirow{2}{*}{\textbf{Model}}} & \multicolumn{3}{c}{\textbf{Attribute Binding}} & \multicolumn{2}{c}{\textbf{Object Relationship}} & \multirow{2}{*}{\textbf{Complex↑}}\\ 
\cline{2-6}
\multicolumn{1}{c}{} & \textbf{Color↑} & \textbf{Shape↑} & \textbf{Texture↑} & \textbf{Spatial↑} &  \textbf{Non-Spatial↑} & \\ 
\hline
LMD \cite{lmd}                   & 0.5150 & 0.5104 & 0.5690 & 0.2772 & 0.2764 & 0.3033\\
LMD+SLD \cite{sld}         & 0.5744 & 0.5054 & 0.5780 & 0.2971 & 0.2748 & 0.2951\\ 
LMD+GenArtist \cite{genartist}               & 0.5640 & 0.5418 & \textbf{0.5956} & 0.3245 & \textbf{0.2825} & 0.3200\\ 
LMD+Ours              & \textbf{0.6493} & \textbf{0.5478} & 0.5950 & \textbf{0.3305} & 0.2804 & \textbf{0.3245}\\ 
\hline
SDXL \cite{sdxl}                  & 0.5673 & 0.4958 & 0.5865 & 0.2035 & 0.3106 & 0.3274\\ 
SDXL+Ours             & \textbf{0.6849} & \textbf{0.5527} & \textbf{0.6412} & \textbf{0.3347} & \textbf{0.3127} & \textbf{0.3502}\\ 
\hline
DALLE-3 \cite{dalle3}              & 0.8011 & 0.6213 & 0.7464 & 0.2272 & 0.3042 & 0.3727\\  
DALLE-3+Ours          & \textbf{0.8341} & \textbf{0.6348} & \textbf{0.7573} & \textbf{0.3080} & \textbf{0.3055} & \textbf{0.3863}\\ 
\hline
\end{tabular}
\end{table*}



\subsection{Implementation Details}

We construct a universal image self-correction platform that leverages OAA and OCS to orchestrate and integrate multiple powerful, task-specific foundation models.
In our experimental setup, we employ LLaMA3-LLaVA-NeXT-8B \cite{llava,improvedllava,llavanext} and LLaMA-3.1-8B \cite{llama3} as MLLM and LLM for the overall process. For attribute editing, we use image editing model UltraEdit \cite{ultraedit}. Also, we set \textit{k}=0.75*\textit{T} to effectively mitigate edge artifacts while preserving essential background details.


\subsection{Benchmarks and Baselines}

We comprehensively demonstrate the effectiveness of Marmot through extensive experiments, particularly highlighting its unified compositional understanding and generation capabilities. We focus our quantitative analysis on the T2I-CompBench benchmark \cite{t2icompbench}. This benchmark focuses on image generation under complex textual prompts involving multiple objects with their respective attributes or interrelationships, serving as a comprehensive evaluation platform for compositional reasoning abilities. 

Since Marmot is a training-free general framework, we compare pre-integration and post-integration metrics within the T2I-CompBench. To systematically evaluate performance improvements, we apply Marmot to state-of-the-art Text-to-Image methods including LMD \cite{lmd}, SDXL \cite{sdxl} and DALL-E 3 \cite{dalle3}. Furthermore, we conduct comparisons by integrating SLD \cite{sld}, GenArtist \cite{genartist}, and Marmot, demonstrating our approach's superior compositional reasoning capability.

To demonstrate the generalizability of our Marmot, we further evaluate SD3.5 \cite{sd3} and FLUX \cite{flux} on T2I-CompBench and SDXL \cite{sdxl} on GenEval \cite{geneval}. 
Beyond text-to-image generation, Marmot transfers seamlessly to layout-to-image tasks such as InstanceDiffusion \cite{instancediffusion} and MIGC \cite{migc}.
Experimental results for this part are presented in the supplementary material.

\subsection{Qualitative evaluation}
As illustrated in Figure~\ref{figure:experiments}, despite the significant advancements in generation quality achieved by existing Text-to-Image models, they still exhibit notable limitations in handling counting, attributes, and spatial relationships. For instance, in the upper example of the figure, SDXL incorrectly generates the flamingo on the left side of the yellow duck, while the attributes ``pink'' and ``yellow'' interfere with the correct generation of the flowers. In contrast, our multi-agent system accurately identifies and localizes these errors, correcting them by invoking different editing tools, ultimately generating images that are highly consistent with the user descriptions. Furthermore, as shown in Figure~\ref{figure:experiments}, our multi-agent system is capable of equally attending to each object in the scene, accurately removing two birds and correcting the color of the deer to white. In comparison, single-agent-based methods such as SLD \cite{sld} and GenArtist \cite{genartist}, even when utilizing advanced GPT-4, struggle to simultaneously handle the positional and attribute information of seven objects. Through multiple experimental validations of SLD and GenArtist, we observe that neither method successfully accounted for the attributes of the deer.

\subsection{Quantitative evaluation}
We integrate the Marmot framework into three text-to-image generation models—LMD \cite{lmd}, SDXL \cite{sdxl}, and DALL-E 3 \cite{dalle3}—and conduct a comparative analysis with GenArtist \cite{genartist} and SLD \cite{sld} on self-correction tasks using images generated by LMD. As shown in Table~\ref{table:evaluation}, Marmot, utilizing only 8B-parameter LLM and MLLM, significantly outperforms GenArtist and SLD in color-related tasks while achieving comparable performance to GenArtist across all other tasks.
Furthermore, for the advanced SDXL and DALL-E 3 models, integrating our framework significantly enhances their performance in attribute binding and object relationships. Specifically, SDXL achieves improvements of 11.76\%, 6.86\%, and 5.47\% in color, shape, and texture tasks, respectively, while DALL-E 3 sees a nearly 3\% boost in both these tasks. In spatial relationship tasks, Marmot also delivers notable performance gains, improving SDXL and DALL-E 3 by 13.12\% and 8.08\%, respectively. These metrics underscore the superiority of the Marmot framework and highlight its potential to open new possibilities in the field of image self-correction.

Additional efficiency experiments, including runtime analysis, are provided in the supplementary material.

\subsection{Ablation Study}

\begin{figure}[t]
   \includegraphics[width=0.47\textwidth]{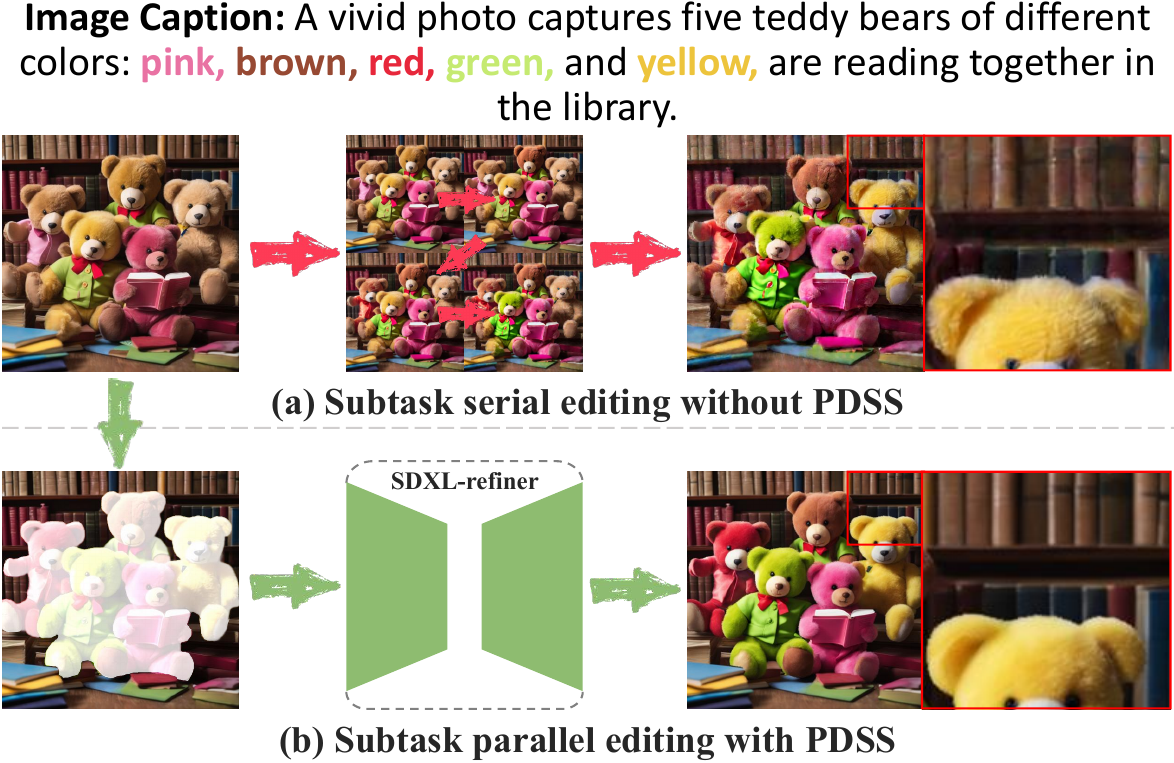}
   \caption{\textbf{Comparison between serial editing and parallel editing with PDSS.} Serial editing distorts backgrounds, but our PDSS parallel editing preserves and enhances them.}
   \label{figure:pdss}
\end{figure}

In this section, we specifically analyze three critical ablation experiments: (1) Parallel Editing Strategy, (2) Decision-Execution-Verification mechanism, and (3) Two-Stage Visual Reasoning Paradigm. The selection rationale for our PDSS base model (SDXL-Refiner), as well as the determination of hyperparameter \textit{K}, are addressed in detail through experimental analysis in the supplementary material.

\textbf{Parallel Editing Strategy.} 
We compare the parallel editing strategy and the serial editing strategy within our proposed PDSS. As shown in Figure~\ref{figure:pdss}, serial editing tends to cause over-editing effects that result in noticeable distortion in background regions. In contrast, our parallel editing framework with PDSS demonstrated in (b) effectively preserves background integrity while simultaneously enhancing object details. These observations underscore the significant role of the PDSS in enhancing the fluidity of the scene. The parallel editing strategy also reduces running time and improves computational efficiency. This will be demonstrated through detailed experiments in the supplementary material.

\textbf{Decision-Execution-Verification mechanism.} 
We fully trust the decisions of the Decision-Maker and eliminate the verification process by Verifier. As shown in Table~\ref{table:ablation}, the introduction of the verification process can improve the attribute binding tasks of color, shape, and texture by 5.10\%, 4.93\%, and 5.19\% respectively, significantly improving the attribute binding ability of text-to-image models. 
This is mainly due to the following reasons: 
\textbf{(1) Performance issues with AI agents.} These agents may produce logical errors, leading to incorrect decisions. For more details, please refer to our supplementary material, where we provide a complete self-correction process, including the logical errors of LLM.
\textbf{(2) Performance issues with editing models.} Even if the Decision-Maker generates a correct decision, the editing model may occasionally fail to produce the desired edited image. Therefore, it is essential to establish a verification mechanism that allows for immediate reversion to the original image and edits with different random seeds. We have also included some error cases in the supplementary material.

\renewcommand{\arraystretch}{1.2}
\setlength{\tabcolsep}{1mm} 
\begin{table}[]
\caption{\textbf{Quantitative Evaluation for Ablation Study.} We remove the verification mechanism and the two-stage visual reasoning paradigm respectively to demonstrate their importance in our framework.}
\label{table:ablation}
\centering
\begin{tabular}{cccc}
\hline
\multicolumn{1}{c}{\multirow{2}{*}{\textbf{Model}}} & \multicolumn{3}{c}{\textbf{Attribute Binding}} \\ 
\cline{2-4} 
\multicolumn{1}{c}{} & \textbf{Color↑} & \textbf{Shape↑} & \textbf{Texture↑} \\ 
\hline
SDXL~\cite{sdxl}                  & 0.5673 & 0.4958 & 0.5865 \\ 
+Ours             & \textbf{0.6849} & \textbf{0.5527} & \textbf{0.6412} \\ 
-w/o verification             & 0.6339 & 0.5034 & 0.5902 \\ 
-w/o two-stage reasoning            & 0.6412 & 0.5421 & 0.6340 \\ 
\hline
\end{tabular}
\end{table}

\textbf{Two-Stage Visual Reasoning Paradigm.} 
We adjust the composition of the Decision-Maker from a collaborative framework of MLLM and LLM to a single MLLM. The object image and object caption are directly fed into the MLLM, which is then queried to determine whether the two are consistent. As shown in Table~\ref{table:ablation}, after removing the two-stage visual reasoning mechanism, the model's performance on color, shape, and texture tasks decreased by 4.37\%, 1.06\%, and 0.72\%, respectively. The results indicate that the drop in the color task is more significant compared to shape and texture tasks. This is primarily because existing image editing models, such as UltraEdit \cite{ultraedit}, are far less capable of editing object shapes and textures than colors. For instance, while the model can easily edit a brown deer's color to yellow when instructed, it often struggles to transform a rectangle bookshelf into a triangular one. This limitation reduces the impact of logical errors in shape and texture tasks. However, the results of the color subtask demonstrate that the two-stage visual reasoning mechanism can effectively reduce logical errors produced by the MLLM, especially when the image editing model performs well, highlighting its advantages more prominently.


\section{Conclusion} \label{sec:conclusion}

In this paper, we propose Marmot, a novel object-level multi-agent framework to address the challenges of accurate object counting, attribute assignment, and spatial relationships in complex multi-object image generation tasks. Marmot leverage an LLM-based Object-Aware Agent, automatically decomposing the self-correction task into fine-grained, object-level subtasks. For each subtasks, we employ a Object Correction Agent with decision-execution-verification mechanism. In this system, Agents operate exclusively on a single object’s segmentation mask or the bounding boxes of object pairs, effectively mitigating inter-object interference and enhancing editing reliability. The introduction of the Pixel-Domain Stitching Smoother further addresses efficiency constraints and distortion issues associated with multi-stage editing pipelines. By leveraging only 8B-parameter open-source MLLMs and LLMs, Marmot achieves superior performance in complex multi-object scenes, outperforming state-of-the-art self-correction frameworks that rely on GPT-4.

\bibliography{aaai2026}


\clearpage

\appendix

\section{Limitations}
\label{sec:limitations}

\begin{figure}[t]
   \includegraphics[width=0.47\textwidth]{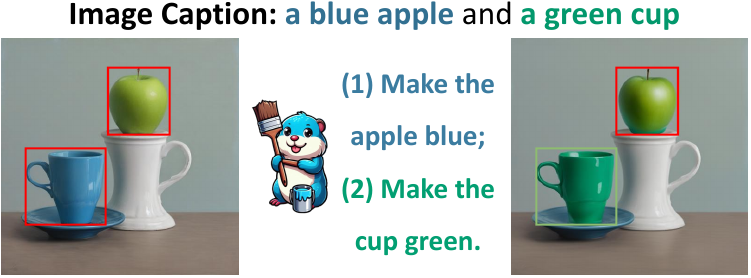}
   \caption{\textbf{The error cases} of Marmot caused by the ability of editing tools.}
   \label{figure:error}
\end{figure}

Although our framework demonstrates superior performance in various aspects, it still has some limitations. 
First, suboptimal performance by any base model can affect the overall quality of the generated images.
As shown in Figure~\ref{figure:error}, the error arises because UltraEdit \cite{ultraedit} fails to edit the green apple to the blue one. However, this issue can be mitigated by the verification mechanism through multiple seed attempts.
Moreover, while Marmot significantly improves generation quality, its multi-agent architecture inevitably introduces additional computational overhead. (1) The decision-execution-verification iterative process, and (2) repeated loading of specialized models on the same GPU collectively contribute to increased inference time. Users can improve Marmot's computational efficiency by (1) reducing the maximum iterations of the decision-execution-verification loop, or (2) employing additional GPUs for base model loading. This demonstrates our framework's inherent flexibility.
The potential misuse of advanced image generation technologies poses risks to social trust and information integrity. Therefore, we strongly advocate for the responsible use of our Marmot to ensure that it is ethically used.

\section{Runtime Efficiency Analysis}
\label{sec:runtime}


 To further assess the practicality of Marmot, we randomly select 50 samples from all prompts in T2I-CompBench \cite{t2icompbench} and conduct end-to-end runtime tests on the SLD \cite{sld}, GenArtist \cite{genartist}, and Marmot, calculating their average post-processing time on a single RTX 3090 GPU (for the parallel editing strategy, the number of GPUs is kept consistent with the number of objects.). Specifically, we use the same images and calculate the time required from inputting the images generated by the baseline model to the output of the corrected images by the self-correcting model. Therefore, the runtime for the LMD model \cite{lmd} is not included in the calculation. Furthermore, we compare the end-to-end runtime of the serial and parallel editing strategies.
 
\begin{figure}[t]
   \includegraphics[width=0.47\textwidth]{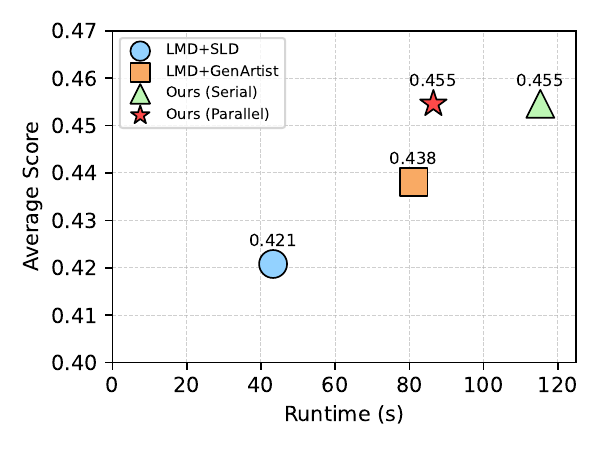}
   \caption{\textbf{Scatter Plot of Model Efficiency vs. Performance:} The x-axis represents runtime, and the y-axis denotes the average score across six tasks in T2I-CompBench.}
   \label{figure:runtime}
\end{figure}

\begin{table}[t]
\centering
\caption{\textbf{Comparison of runtime between serial editing and parallel editing} on T2I-CompBench.}
\footnotesize
\begin{tabular}{cc}
\hline
\textbf{Model}      & \textbf{Runtime↓} \\ \hline
LMD \cite{lmd}             & -                 \\
+SLD \cite{sld}            & \textbf{43.27 s}  \\
+GenArtist \cite{genartist}      & 81.16 s           \\
+Ours (Serial)   & 115.31 s          \\ 
+Ours (Parallel) & 86.47 s           \\ \hline
\end{tabular}
\label{table:runtime}
\end{table}

 As illustrated in Figure~\ref{figure:runtime} and Table~\ref{table:runtime}, Marmot achieves a runtime that is approximately $2 \times$ longer than SLD while remaining comparable to that of GenArtist. This is primarily attributed to the multi-round decision-making process introduced by the verification mechanism and the overhead of repeatedly loading LLM and MLLM onto the same RTX 3090 GPU when processing identical objects.
 Notably, the results in Table~\ref{table:runtime} show that serial editing takes 115.31 seconds on average, while parallel editing reduces this to 86.47 seconds, showing a performance improvement of approximately 25\%. Since T2I-CompBench consists primarily of prompts involving two objects, this improvement is not particularly significant. In scenarios with a larger number of objects, our parallel editing strategy by PDSS will demonstrate greater superiority.


\section{Additional Ablation study}
\label{sec:additional ablation}

\begin{figure}[b]
   \includegraphics[width=0.47\textwidth]{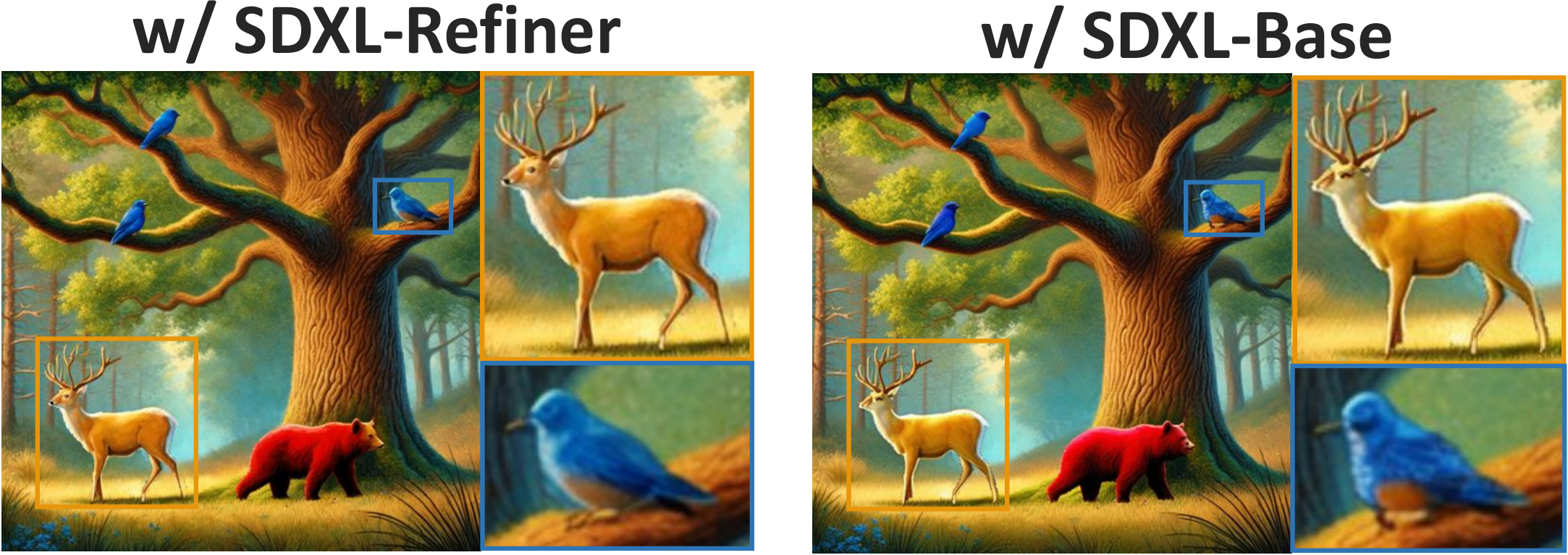}
   \caption{\textbf{Comparison between different base model of PDSS.} SDXL-Refiner further refines the details of objects within the image.}
   \label{figure:refiner}
\end{figure}

\begin{figure*}[t]
	\centering
	\includegraphics[width=1\textwidth]{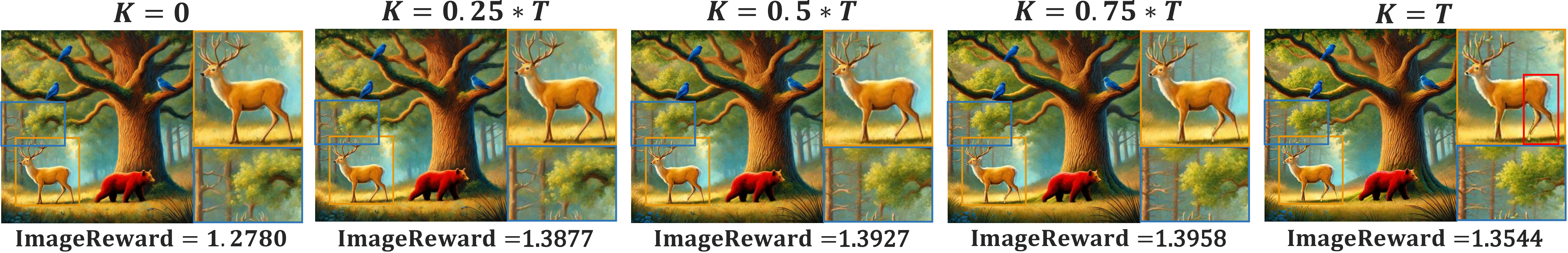} %
	\caption{\textbf{Ablation study on the value of K steps.} As the value of K decreases, the edges of objects become increasingly smooth. However, this comes at the cost of deteriorating detail representation.}
	\label{figure:k}
\end{figure*}

\textbf{Base model of PDSS.}
To further investigate the influence of different base models, we change the base model of our PDSS from SDXL-Refiner to SDXL-Base \cite{sdxl}. As depicted in Figure~\ref{figure:refiner}, the SDXL-Base model is more inclined to reconstruct composite images. Conversely, SDXL-Refiner further refines the details of objects in the image, and the overall aesthetic appeal and realism of the image are significantly enhanced. Current text-to-image models and image editing models often struggle to generate or edit small objects, frequently resulting in outputs that lack detail and appear unrealistic. By utilizing SDXL-Refiner as the base model of PDSS, we are able to mitigate these performance issues mentioned above. 

\textbf{\textit{K} for Optimal Aesthetic Balance in PDSS.} 
Our proposed PDSS has the hyperparameter \textit{K} to control the switching timesteps between the background detail preservation stage and the object edge smoothing stage.
We select different values of \textit{K} to qualitatively assess the impact of \textit{K} on PDSS. As illustrated in Figure~\ref{figure:k}, distinct edge artifacts emerge on the deer's legs when \textit{K}=\textit{T}. Conversely, when \textit{K} is reduced, the background becomes excessively smooth, which is reflected in the increasingly poor detail of the tree branches within the blue border. To balance these two factors, we utilize the ImageReward model \cite{imagereward} to evaluate the aesthetic score of each image. Our findings demonstrate that setting \textit{K} to 0.75*\textit{T} achieves an optimal balance - effectively mitigating edge artifacts while preserving essential background details.


\begin{figure}[b]
    \centering
    \includegraphics[width=0.47\textwidth]{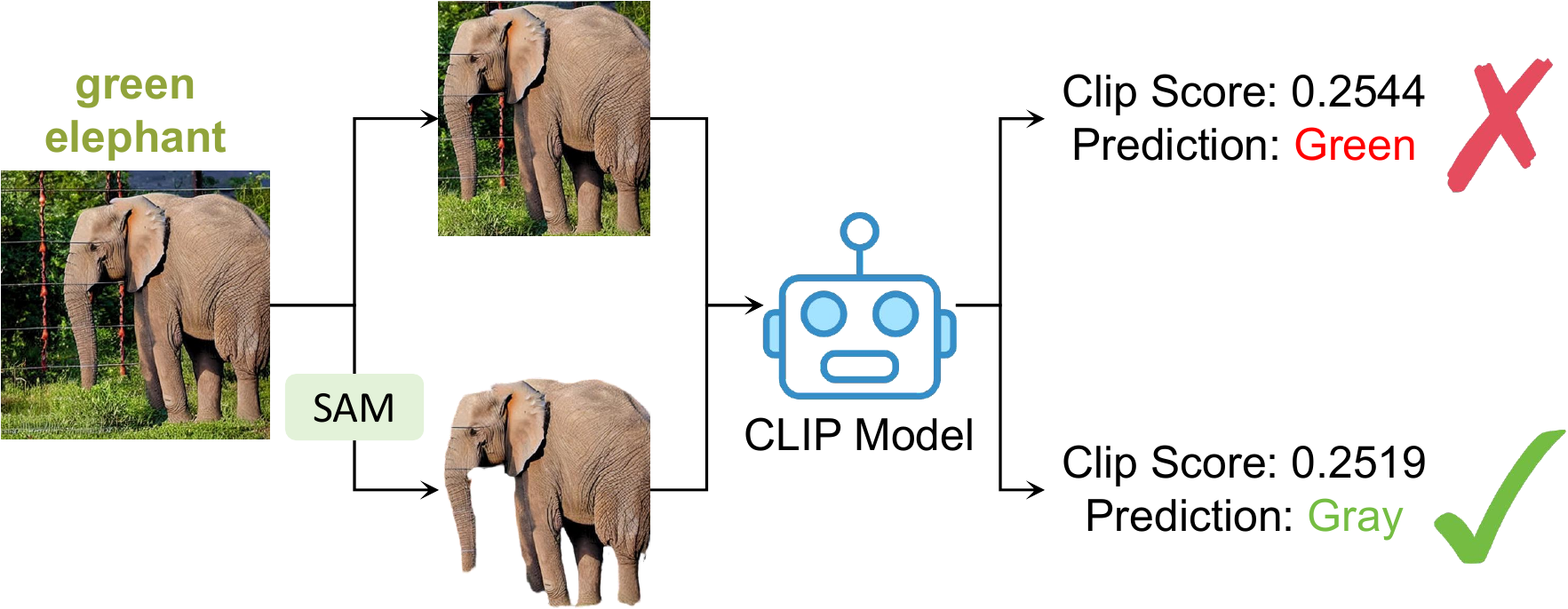}
    \caption{\textbf{Metrics Comparisons between InstanceDiffusion and Ours.} We isolate the object using a segmentation model to eliminate environmental influences.}
    \label{figure:eval}
\end{figure}

\begin{figure}[t]
    \centering
    \includegraphics[width=0.47\textwidth]{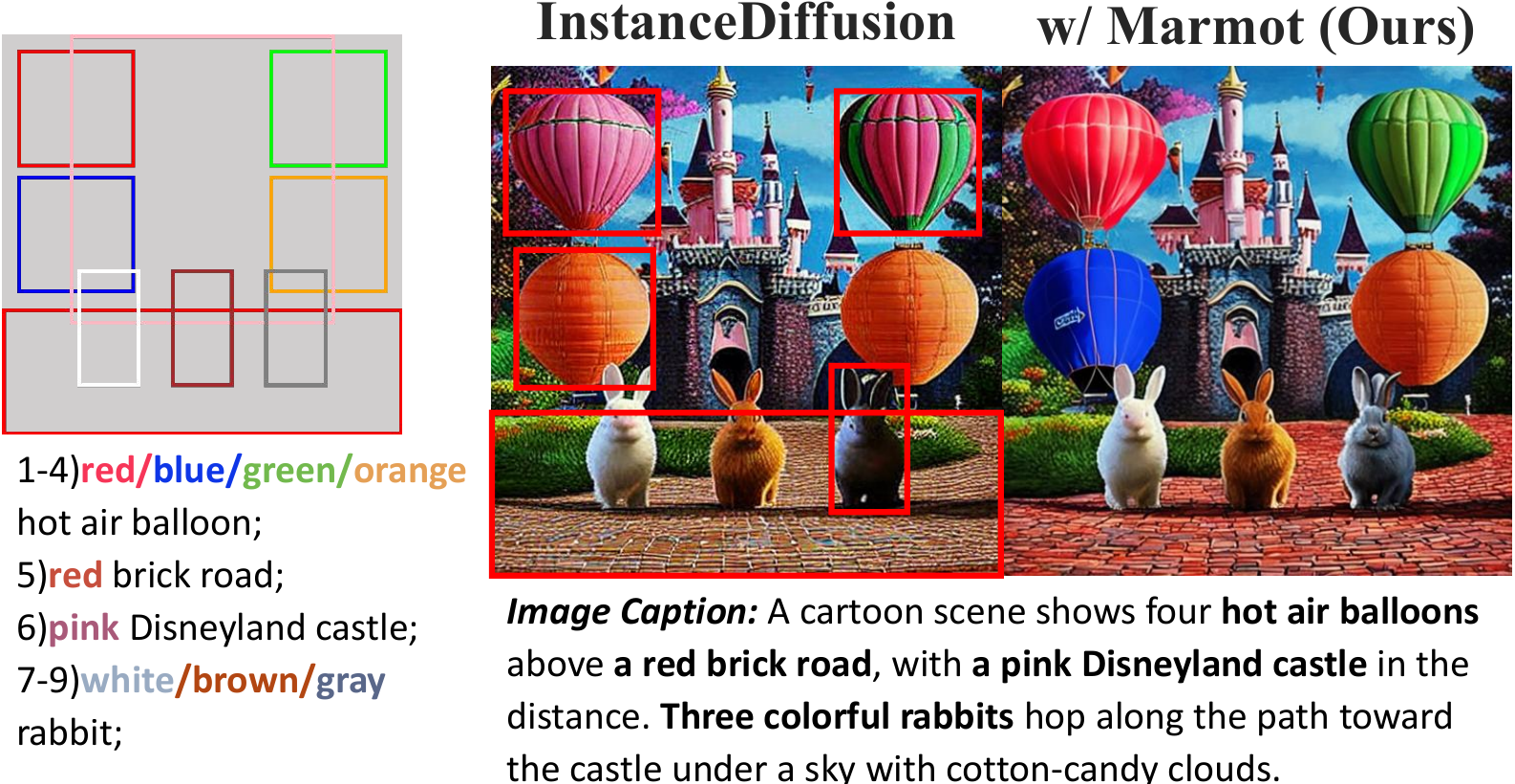}
    \caption{\textbf{Comparisons with InstanceDiffusion.} We individually edited the colors of the balloons, ensuring that the colors of different objects would not interfere with each other.}
    \label{figure:instdiff}
\end{figure}

\renewcommand{\arraystretch}{1.2}
\setlength{\tabcolsep}{1mm} 
\begin{table}[t]
\caption{\textbf{Evaluation for Layout-to-Image generation.}``Acc'' represents attribute accuracy, ``Clip'' denotes the local Clip score.}
\label{table:Layout-to-Image}
\centering
\begin{tabular}{ccccc}
\hline
\multicolumn{1}{c}{\multirow{2}{*}{\textbf{Model}}} & \multicolumn{2}{c}{\textbf{Color}} & \multicolumn{2}{c}{\textbf{Texture}} \\ 
\cline{2-5} 
\multicolumn{1}{c}{} & \textbf{Acc↑} & \textbf{Clip↑} & \textbf{Acc↑} & \textbf{Clip↑} \\ 
\hline
InstanceDiffusion                 & 51.66\% & 0.2445 & 19.50\% & 0.2156 \\ 
+Ours            & \textbf{57.90\%} & \textbf{0.2550} &   \textbf{24.95\%} &
  \textbf{0.2198} \\ \hline
MIGC             & 57.50\% & 0.2511 & 16.79\%   & 0.2071 \\ 
+Ours             & \textbf{61.37\%} &
  \textbf{0.2563} & \textbf{21.95\%} &
  \textbf{0.2152} \\ 
\hline
\end{tabular}
\end{table}

\section{Unified Text-to-Image and Layout-to-Image Generation}
\label{sec:unified}
Additionally, our Marmot can be seamlessly transferred to the task of layout-to-image generation \cite{gligen,migc,instancediffusion}. In this application scenario, since the bounding boxes and object information are provided by the user, only the OCS for attribute self-correction and the PDSS module need to be activated to achieve effective adaptation and performance optimization of the model in this field. This is equivalent to decomposing the subtasks and extracting the layout by user themselves, thereby skipping the process of OAA and counting self-correction.

\renewcommand{\arraystretch}{1.2}
\begin{table*}[h]
\caption{\textbf{More Quantitative Results on GenEval.}}
\label{table:geneval}
\centering
\begin{tabular}{cccccccc}
\hline
\textbf{Model} & \textbf{Overall} & \textbf{Single Object} & \textbf{Two Object} &\textbf{Counting} &\textbf{Colors} &\textbf{Position} &\textbf{Color Attribution} \\ 
\hline
SDXL \cite{sdxl} & 0.55 & \textbf{0.99} & 0.76 & 0.38 & 0.89 & 0.08 & 0.21 \\ 
+Ours  & \textbf{0.66} & \textbf{0.99} & \textbf{0.82} & \textbf{0.63} & \textbf{0.92} & \textbf{0.22} & \textbf{0.35} \\  \hline
\end{tabular}
\end{table*}

\renewcommand{\arraystretch}{1.2}
\begin{table*}[h]
\caption{\textbf{More Quantitative Results on T2I-CompBench.}}
\label{table:stronger}
\centering
\begin{tabular}{ccccccc}
\hline
\multicolumn{1}{c}{\multirow{2}{*}{\textbf{Model}}} & \multicolumn{3}{c}{\textbf{Attribute Binding}} & \multicolumn{2}{c}{\textbf{Object Relationship}} & \multirow{2}{*}{\textbf{Complex↑}}\\ 
\cline{2-6}
\multicolumn{1}{c}{} & \textbf{Color↑} & \textbf{Shape↑} & \textbf{Texture↑} & \textbf{Spatial↑} &  \textbf{Non-Spatial↑} & \\ 
\hline
FLUX.1-dev \cite{flux}                  & 0.7479 & 0.4645 & 0.6211 & 0.2790 & 0.3067 & 0.3649 \\ 
FLUX.1-dev+Ours             & \textbf{0.7776} & \textbf{0.5247} & \textbf{0.6735} & \textbf{0.3430} & \textbf{0.3085} & \textbf{0.3769}\\ \hline
Stable Diffusion 3.5-medium \cite{sd3}  & 0.7985 & 0.5770 & 0.7200 & 0.2989 & 0.3149 & 0.3637 \\ 
Stable Diffusion 3.5-medium+Ours        & \textbf{0.8320} & \textbf{0.6128} & \textbf{0.7379} & \textbf{0.3803} & \textbf{0.3166} & \textbf{0.3737}\\ 
\hline
\end{tabular}
\end{table*}

\subsection{Benchmark, Baselines, and Evaluation Metrics}
For the Layout-to-Image generation tasks, we randomly sample 100 layouts in COCO \cite{coco} and assign a color to each object while maintaining the original layout. Similarly, we sample another 100 layouts for texture attributes. For instance in Figure~\ref{figure:eval}, we insert a color ``green'' for ``elephant'' and provide it as input to the layout-to-image model, hoping the model can generate a green elephant.

In terms of the baselines, we analyze pre-integration and post-integration metrics within the COCO datasets. We employ Marmot to state-of-the-art Layout-to-Image methods, such as InstanceDiffusion \cite{instancediffusion} and MIGC \cite{migc}, to evaluate the performance of Marmot.

As for the metrics, we employ an InstanceDiffusion \cite{instancediffusion} method to isolate image objects for precise attribute evaluation. Furthermore, we ensure a pure focus on the objects by using the SAM model \cite{sam} to reduce the influence of environmental attributes on the evaluation process. These isolated objects are then provided to the CLIP model \cite{clip} to calculate the local CLIP Score and predict their attributes (the attribute that has the highest similarity to the attributes we provided), measuring the prediction accuracy against the attributes specified in the object captions. As depicted in Figure~\ref{figure:eval}, the input description specifies a ``green elephant''. However, InstanceDiffusion generates an image featuring a gray elephant with a green background. If we directly use the CLIP model to predict the color of the cropped elephant image, the output would still be ``Green'', due to the dominant green background. To mitigate this environmental interference, we employ the SAM model \cite{sam} to segment and isolate the elephant object. The resulting color prediction then accurately reflects the true color of the elephant as ``Gray''.

\subsection{Qualitative evaluation}
For the Layout-to-Image tasks, we individually edit the colors of the balloons and turn the color of the road red, ensuring that the colors of different objects would not interfere with each other, as shown in Figure~\ref{figure:instdiff}. After undergoing the OCS process of Marmot, the consistency between the generated images and their local descriptions can be significantly enhanced.

\subsection{Quantitative evaluation}
For the Layout-to-Image generation tasks in Table~\ref{table:Layout-to-Image}, we help InstanceDiffusion and MIGC achieve precision improvements of almost 5\% in the COCO dataset. Notably, InstanceDiffusion achieves an accuracy of 51.66\% and a CLIP score of 0.2445 for color, which improves to 57.90\% and 0.2550 with our method. MIGC shows an accuracy of 57.50\% and a CLIP score of 0.2511, which increases to 61.37\% and 0.2563 with our enhancement. In terms of texture, InstanceDiffusion's accuracy is 19.50\% with a CLIP score of 0.2156, improving to 24.95\% and 0.2198. MIGC's texture accuracy is 16.79\% with a CLIP score of 0.2071, which rises to 21.95\% and 0.2152 with our contribution. However, the assistance provided by our Marmot is less significant compared to text-to-image models. This is primarily because, while layout-to-image models can effectively construct scenes, they often lose objects. In such cases, we cannot correctly modify attributes without a target. In future work, we will consider how to address the issue of object loss.

\section{Additional Quantitative Results}
\label{sec:stronger}
\subsection{Comparisons on the GenEval benchmark}
To further validate the generalization capability of our framework, we conduct additional evaluations of SDXL \cite{sdxl} on the GenEval benchmark \cite{geneval}. GenEval is an object-oriented, fine-grained evaluation framework that leverages detection models to automatically assess text-to-image models on compositional attributes such as object position, count, and color, demonstrating high consistency with human evaluations. As shown in Table~\ref{table:geneval}, integrating our framework significantly improves SDXL's performance in counting, position, and attribute binding, with gains of 0.25, 0.14, and 0.14, respectively. The overall average score increases by 0.11, underscoring the superiority and strong generalization ability of our framework.

\begin{figure*}[t]
	\centering
	\includegraphics[width=0.7\textwidth]{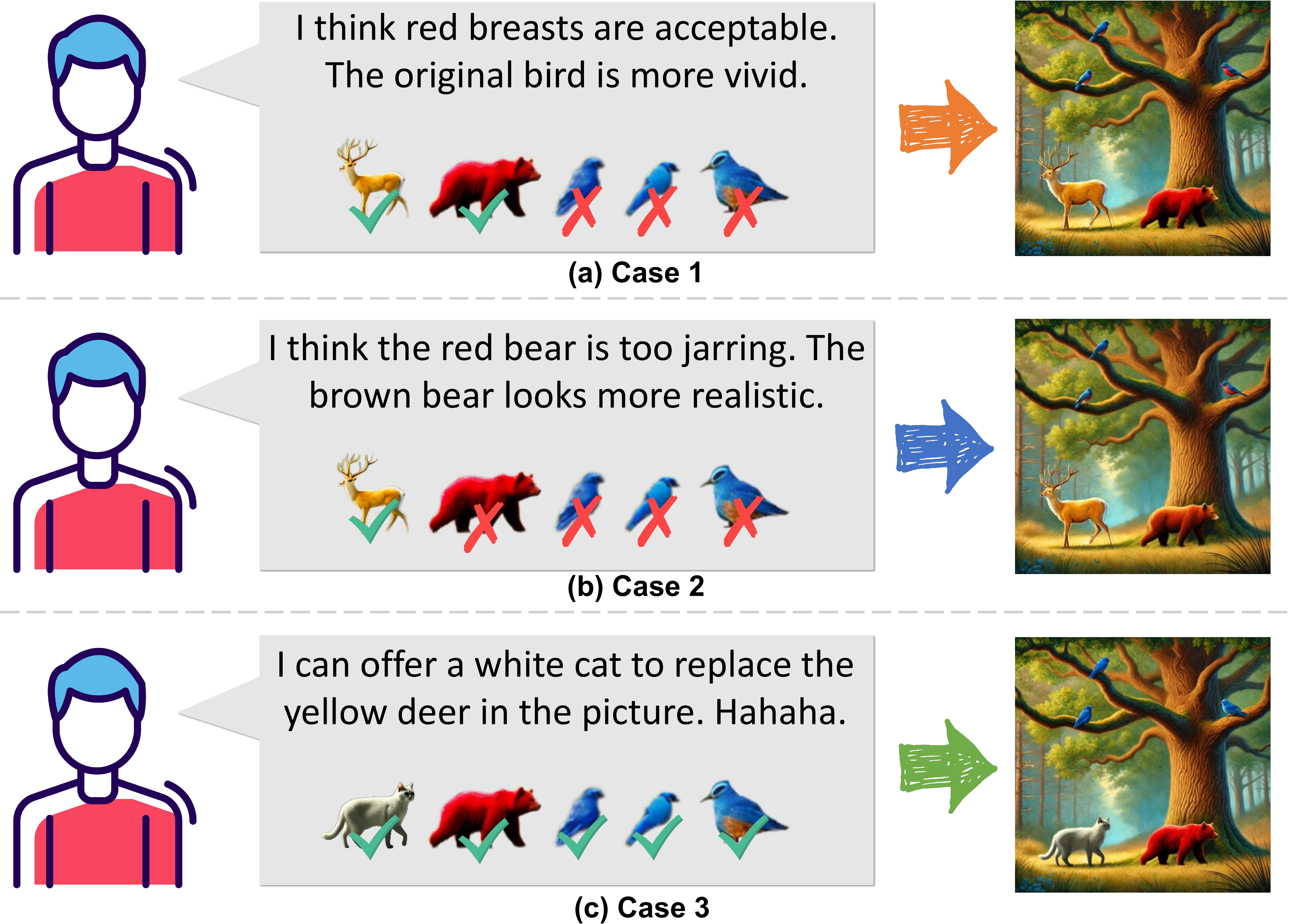} %
	\caption{\textbf{Examples of interactive object correction with user feedback.} Through user preference-based adjustments, Marmot achieves higher levels of personalized image self-correcting.}
	\label{figure:user}
\end{figure*}

\subsection{Comparison with Stronger Baselines}
Recently, FLUX \cite{flux} and SD3.5 \cite{sd3} achieve new state-of-the-art results by adopting the Diffusion Transformer \cite{dit} as their base structure and using flow matching techniques for training. This leads to improved stability and sample efficiency in large-scale diffusion models. To more precisely situate our proposed approach within the current state-of-the-art, we meticulously conduct a comprehensive comparison on the T2I-CompBench benchmark \cite{t2icompbench} between stronger baselines and our Marmot. As shown in Table~\ref{table:stronger}, our experimental results demonstrate consistent performance improvements for both FLUX and SD3.5 across all tasks. In the attribute binding task, FLUX achieves significant performance gains of 2.97\%, 6.02\%, and 5.24\% across different metrics, while SD3.5 shows improvements of 3.35\%, 3.58\%, and 1.79\%, respectively. For the object spatial relationship task, FLUX and SD3.5 demonstrate performance improvements of 6.4\% and 8.14\%. These results not only validate the reliability of Marmot but also highlight its adaptability to diverse model architectures.

\section{Interactive Image Refinement via User Evaluation and Feedback}
\label{sec:user}
Incorporating a human-interactive process into image self-correction can improve alignment with human preferences. As illustrated in Figure~\ref{figure:user}, our framework allows users to: (1) select preferred objects for seamless composition into the target image, and (2) manually edit objects before integration into the Marmot pipeline. These capabilities demonstrate the framework's transparency and controllability in accommodating diverse user inputs while maintaining robust generation quality. The interactive nature of this approach enables precise preference tuning, leading to more satisfactory output that better matches user expectations. This human-interactive process is also a means to address suboptimal performance by base models.

\section{Additional Qualitative Results}
\label{sec:additional}
\begin{figure*}[h]
	\centering
	\includegraphics[width=1\textwidth]{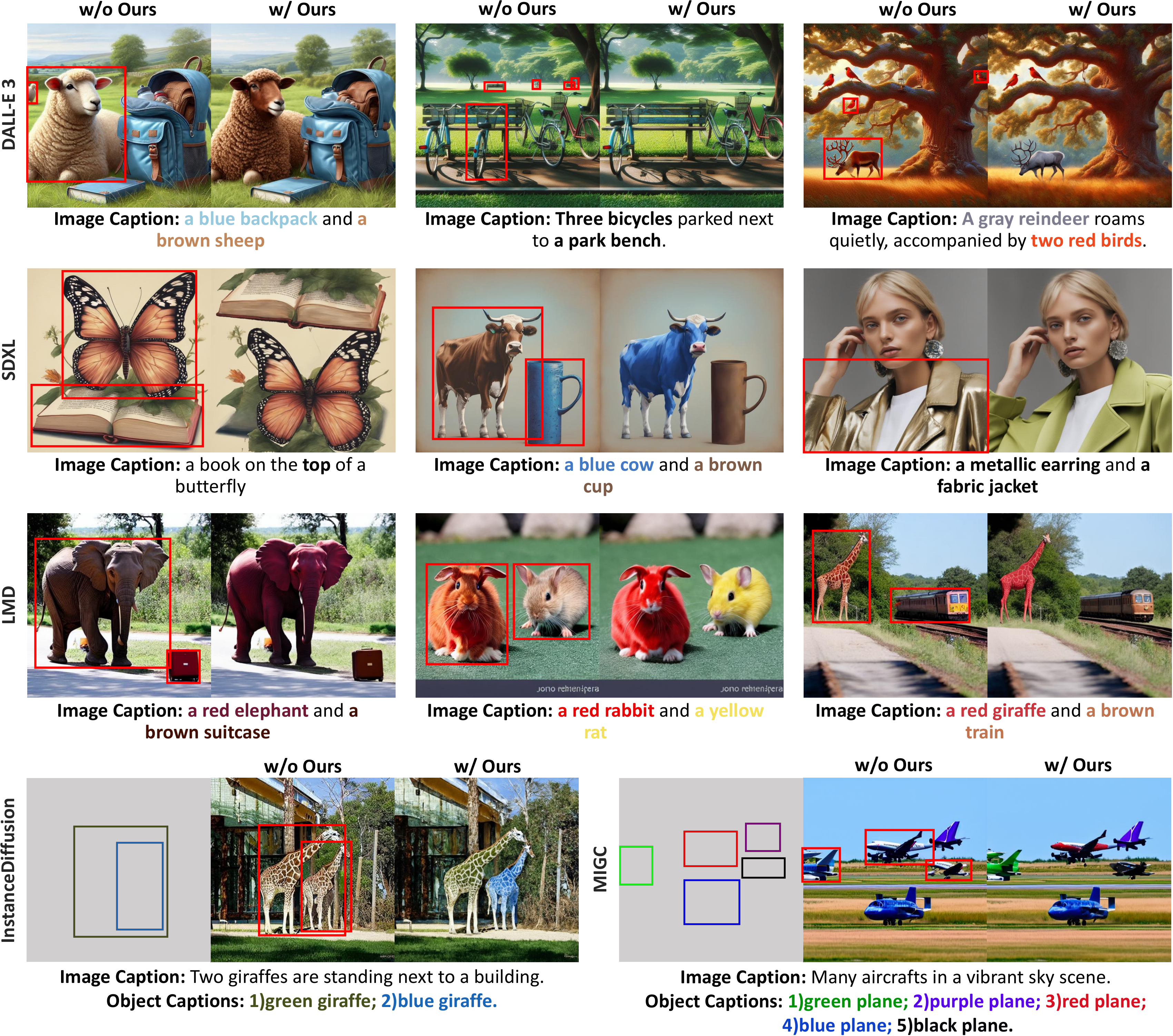} %
	\caption{\textbf{More Qualitative Results on Text-to-Image tasks and Layout-to-Image tasks.} In each pair of images, the left side represents the original image, while the right side depicts the output following the integration of our Marmot model.}
	\label{figure:supp}
    \vspace{-0.3cm}
\end{figure*}

Figure~\ref{figure:supp} displays additional outcomes for the Text-to-Image and Layout-to-Image models. These figures offer a comprehensive visual comparison that elucidates the performance disparities between state-of-the-art models prior to and following the incorporation of our Marmot framework. In each pair of images, the left side represents the original image, while the right side depicts the output following the integration of our Marmot model. Our Marmot framework enhances previous state-of-the-art models by substantially improving the precision among three critical dimensions (counting, attributes, and spatial relationships), thereby achieving image generation that closely aligns with user descriptions.

\section{Overall Process}
\label{sec:overall}
\begin{figure*}[ht]
	\centering
	\includegraphics[width=1\textwidth]{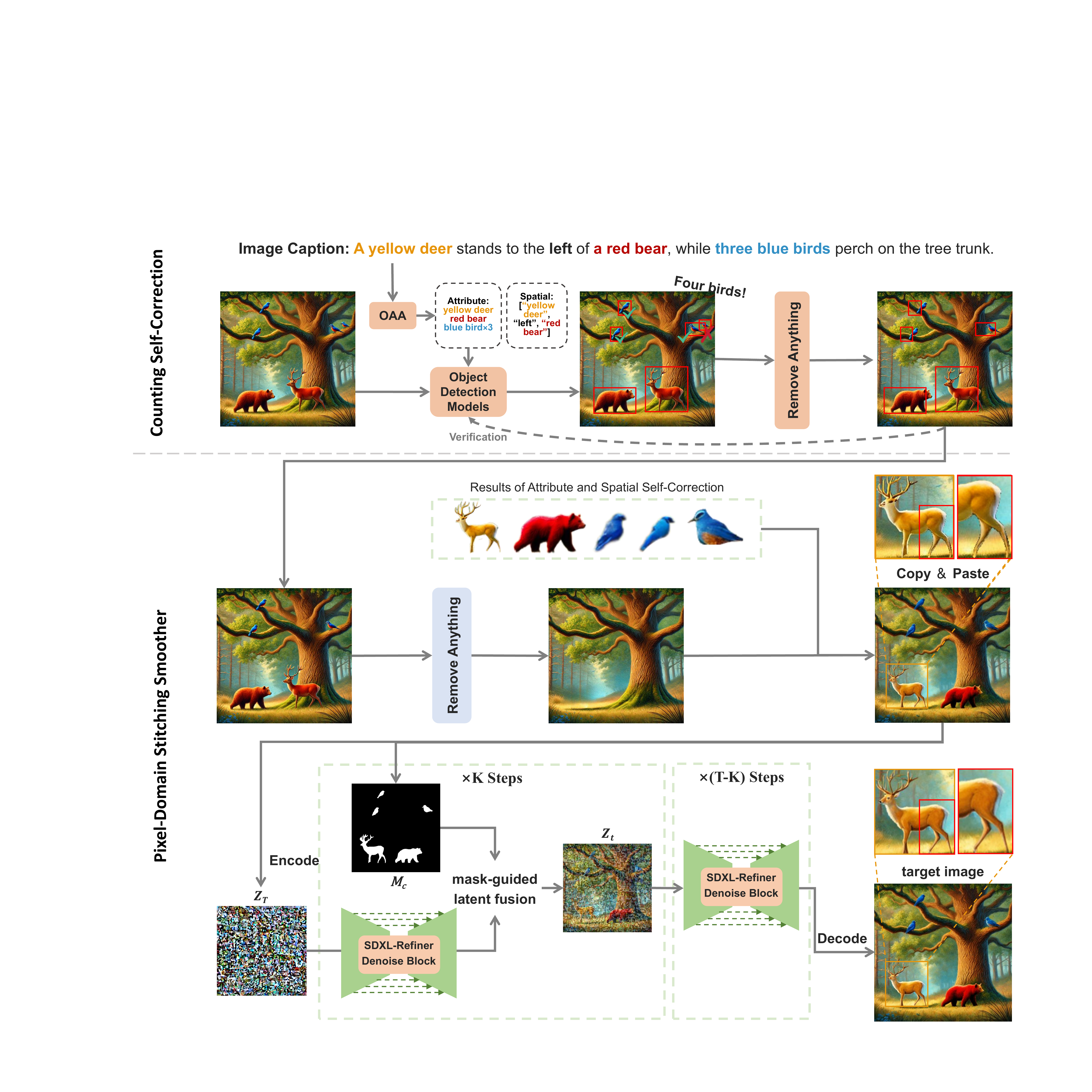} %
	\caption{\textbf{The detailed record of the overall process of a self-correction.}}
	\label{figure:overall}
\end{figure*}

In this section, we provide a detailed record of the overall process of self-correction. Figure~\ref{figure:overall} illustrates the complete steps from image input to the final output, including key stages such as layout extraction, decision-making, editing, verification, and Pixel-Domain Stitching Smoother (PDSS). Through this process, we can clearly see how each stage works together to achieve precise correction and optimization of the image.

Firstly, we employ an LLM-based OAA to decompose self-correction task. 
Then, we conduct counting self-correction to extract the layout and local descriptions. 
The manuscript provides comprehensive explanations of both attribute and spatial self-correction, accompanied by complete prompts and responses from LLM and MLLM. Therefore, these concepts will not be reiterated here. 
Subsequently, we provide a detailed description of our proposed PDSS. As shown in Figure~\ref{figure:overall}, PDSS begins with the removal of all objects that need to be modified, utilizing SAM \cite{sam} and LAMA \cite{lama}. The edited object is then directly pasted into the modified image. This approach allows us to address each subtask parallelly, ensuring a modular and efficient workflow. Finally, we leverage SDXL-Refiner \cite{sdxl} and mask-guided two stage latent space optimization to smooth the edge of composite objects and enhance the detail of objects. 

\begin{figure*}[ht]
	\centering
	\includegraphics[width=1\textwidth]{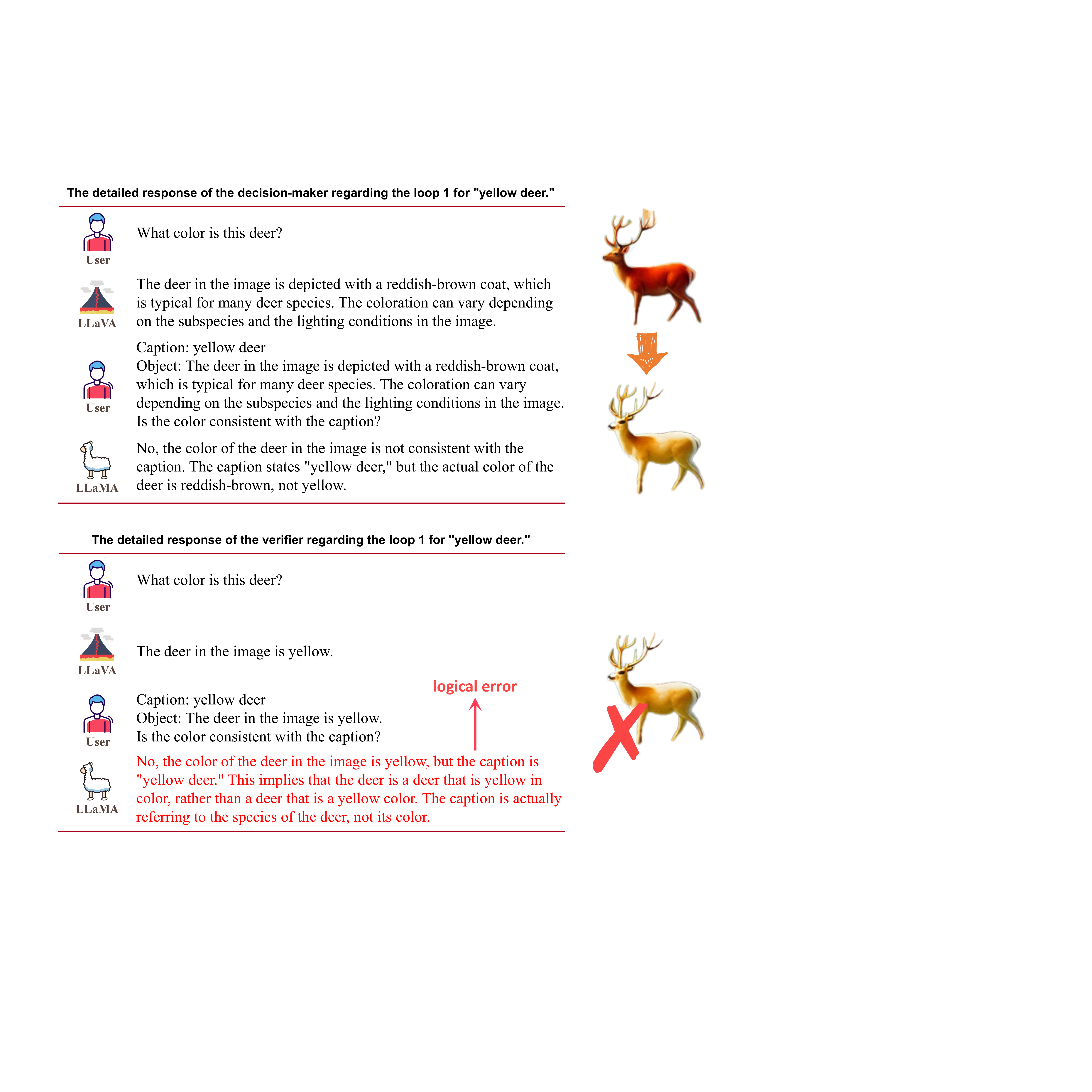} %
	\caption{\textbf{The detailed response of the decision-maker and verifier regarding the loop 1 for "yellow deer".} In this case, the logical error of the LLM occurred, which proved the effectiveness of our verification mechanism.}
	\label{figure:deer1}
    \vspace{-0.5cm}
\end{figure*}

\begin{figure*}[ht]
	\centering
	\includegraphics[width=1\textwidth]{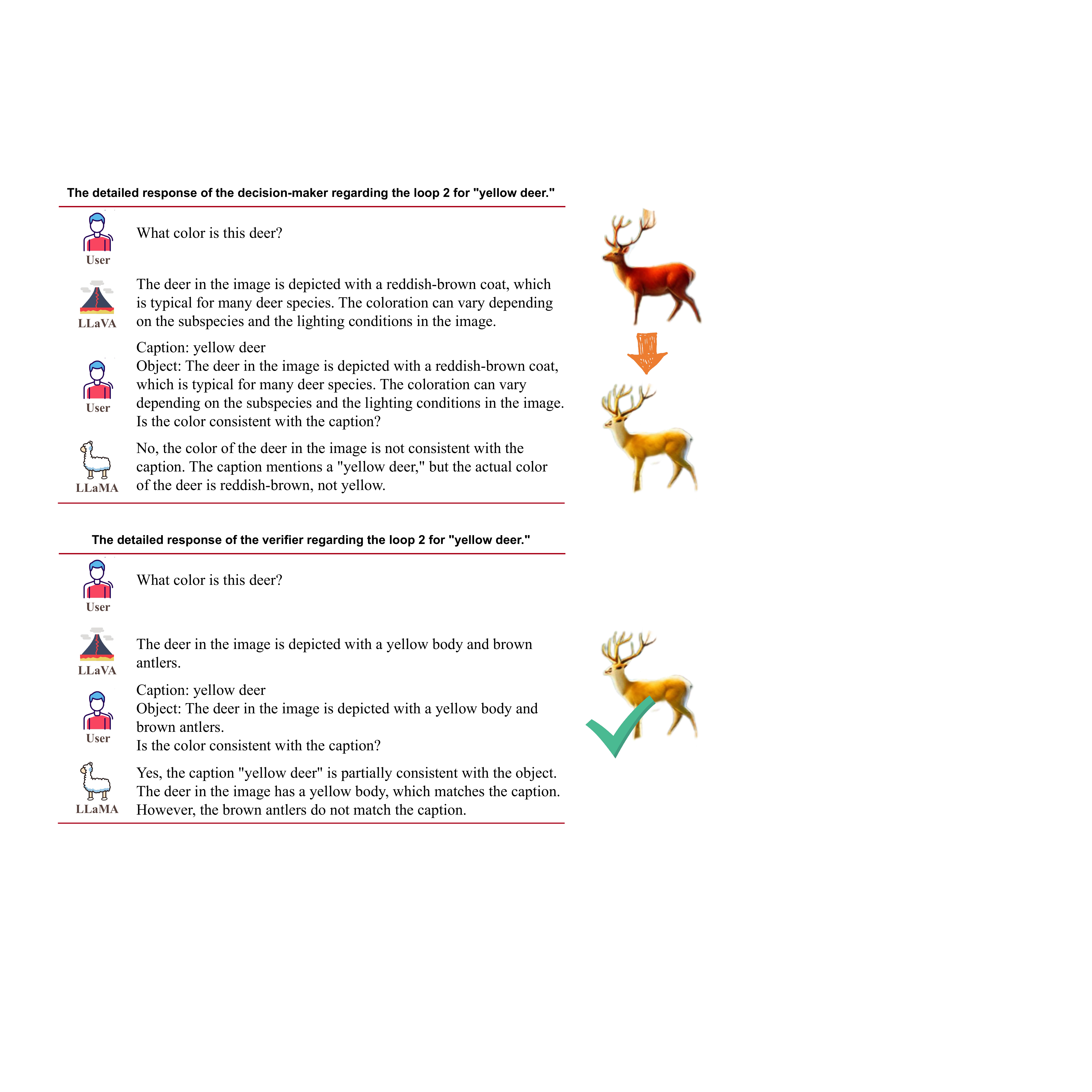} %
	\caption{\textbf{The detailed response of the decision-maker and verifier regarding the loop 2 for "yellow deer".}}
	\label{figure:deer2}
\end{figure*}

\begin{figure*}[ht]
	\centering
	\includegraphics[width=1\textwidth]{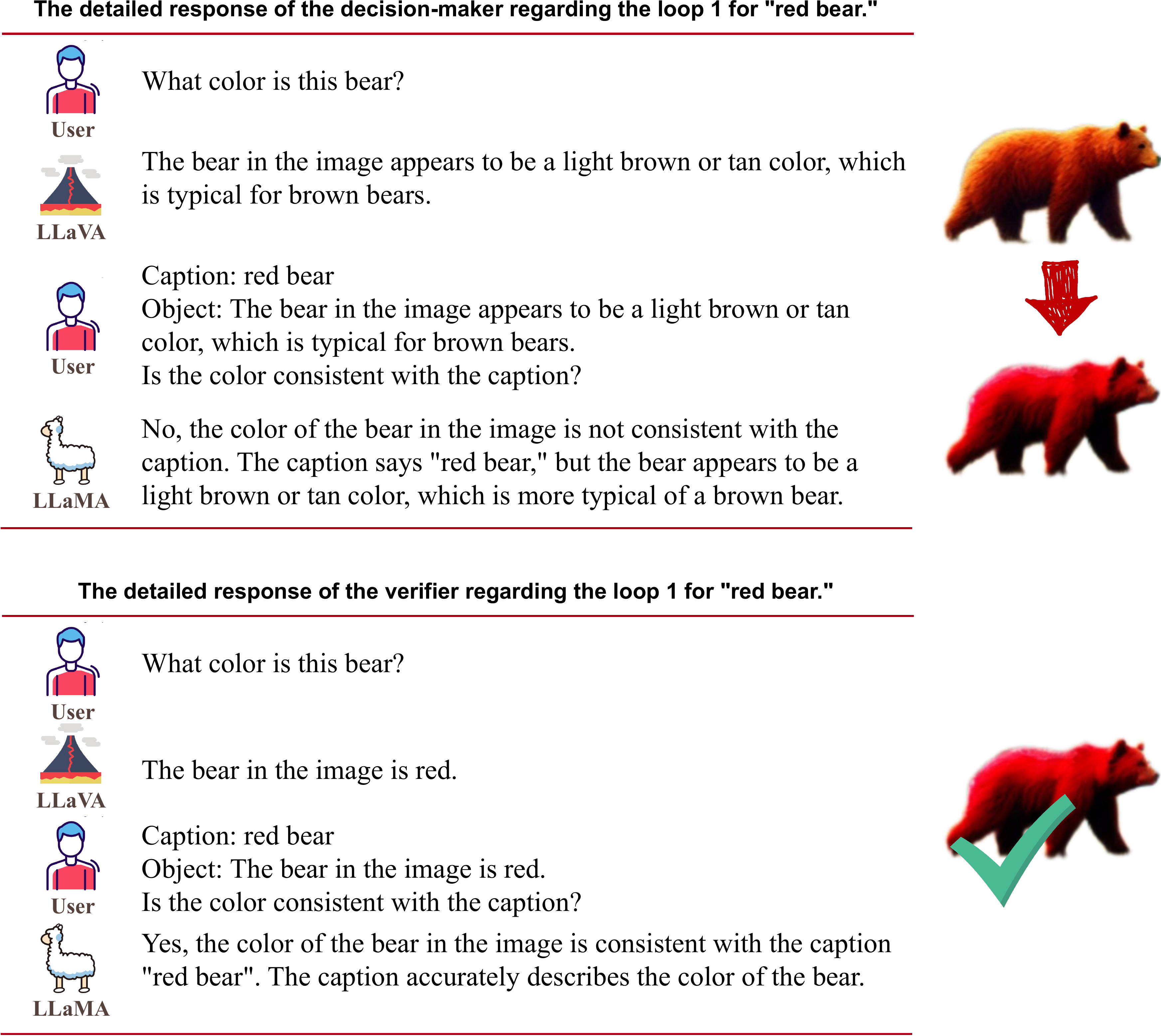} %
	\caption{\textbf{The detailed response of the decision-maker and verifier regarding the loop 1 for "red bear".}}
	\label{figure:bear1}
\end{figure*}

\begin{figure*}[ht]
	\centering
	\includegraphics[width=1\textwidth]{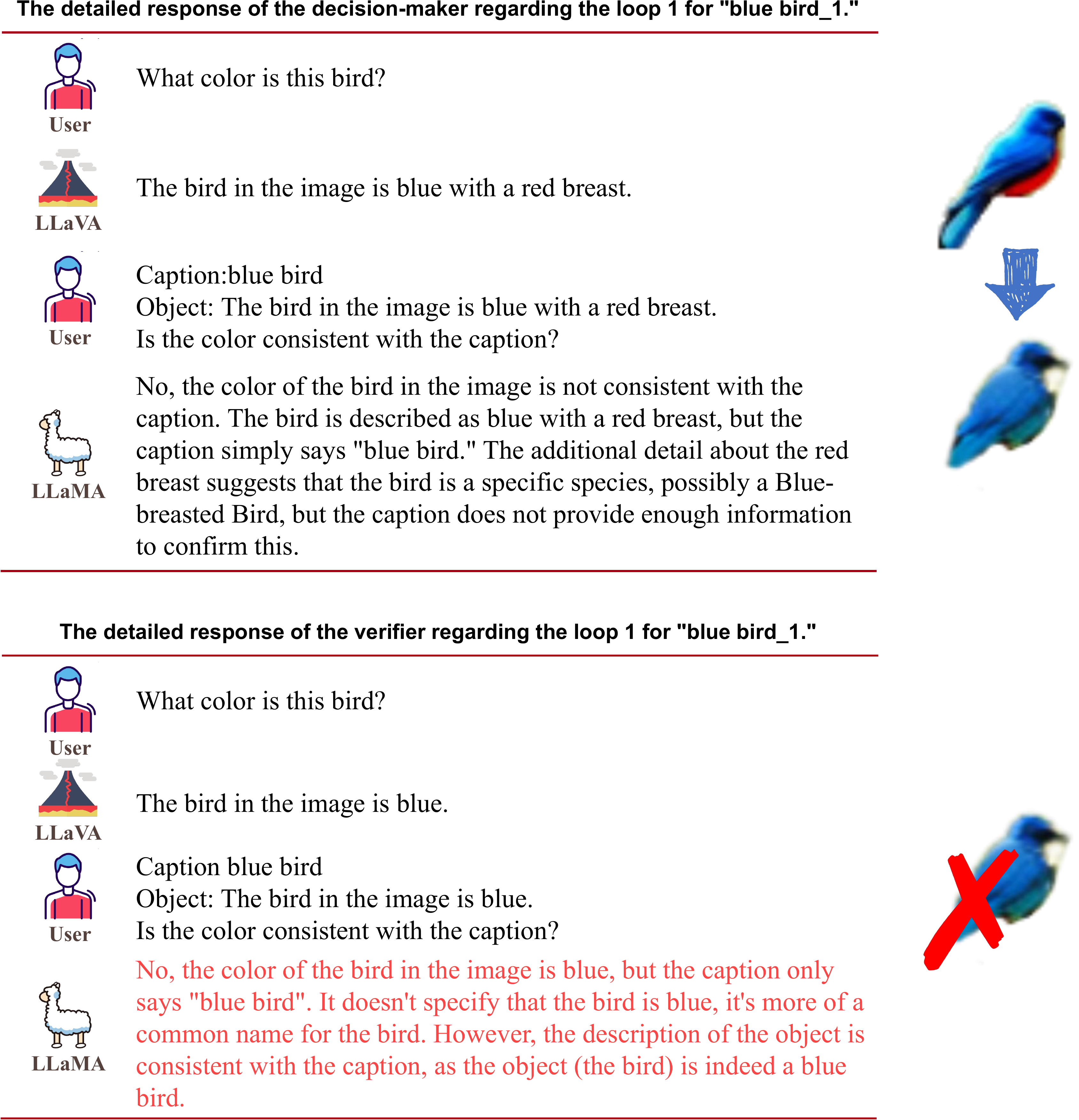} %
	\caption{\textbf{The detailed response of the decision-maker and verifier regarding the loop 1 for "blue bird".}}
	\label{figure:bird1}
\end{figure*}

\begin{figure*}[ht]
	\centering
	\includegraphics[width=1\textwidth]{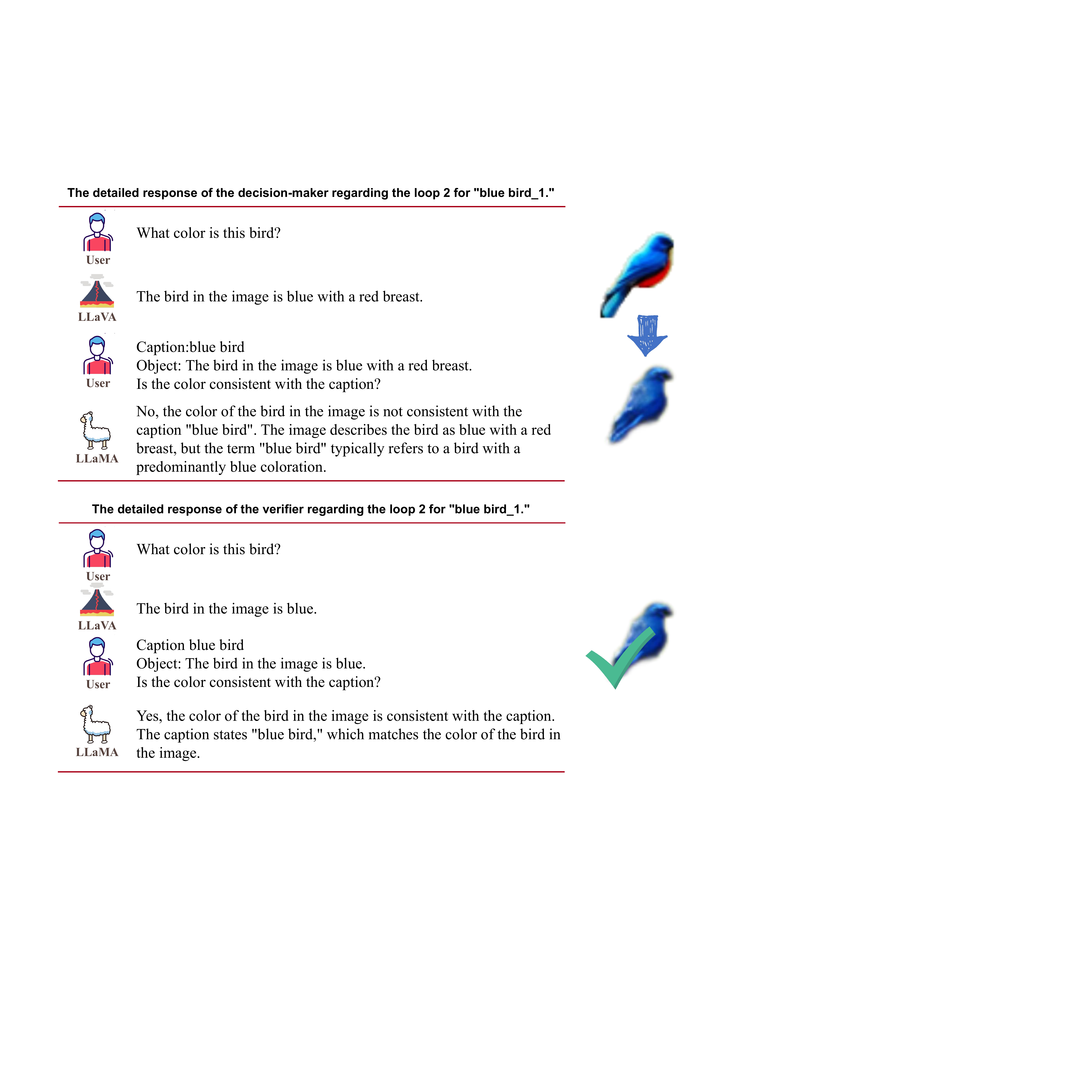} %
	\caption{\textbf{The detailed response of the decision-maker and verifier regarding the loop 2 for "blue bird".}}
	\label{figure:bird2}
\end{figure*}

\begin{figure*}[h]
	\centering
	\includegraphics[width=1\textwidth]{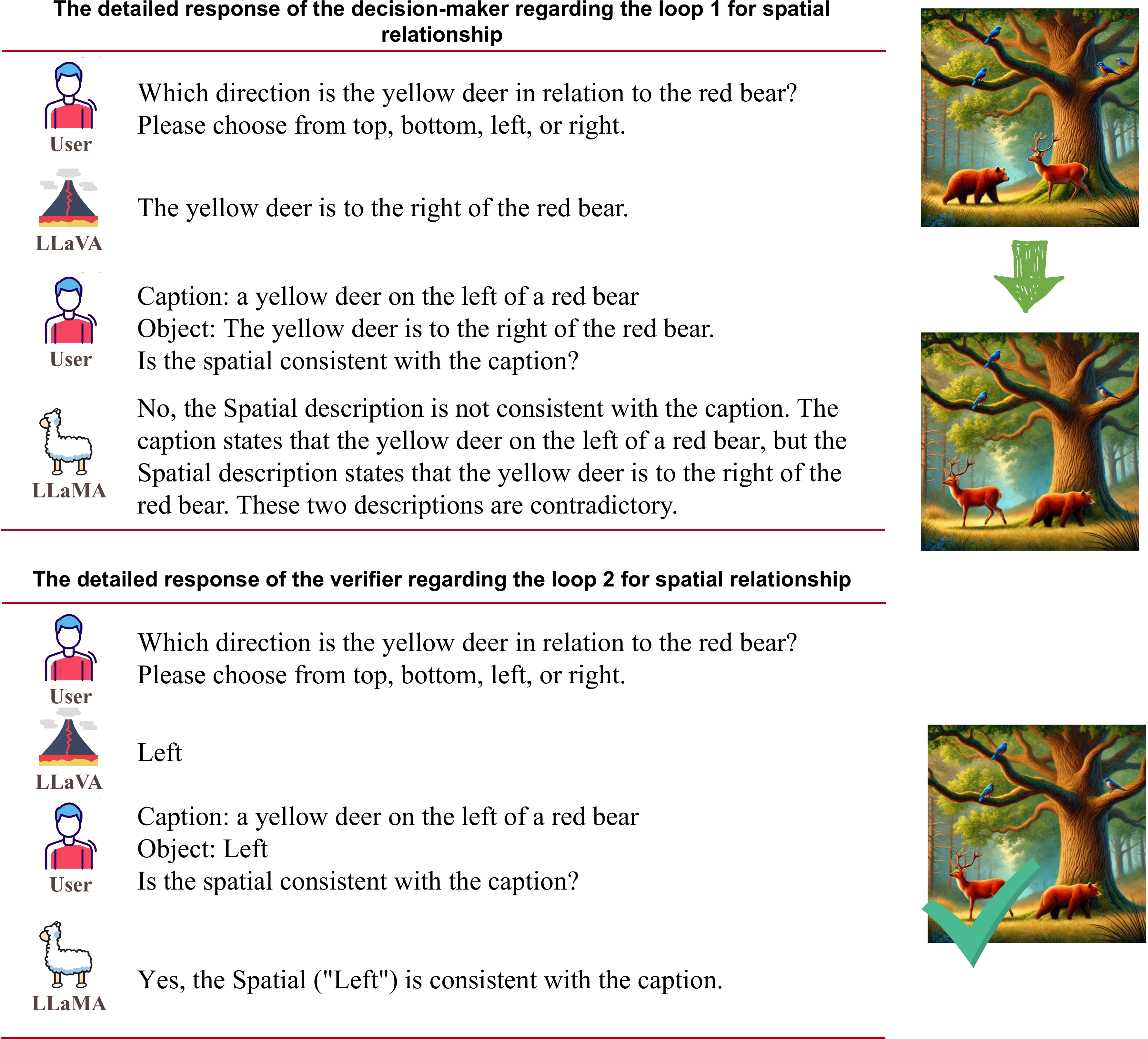} %
	\caption{\textbf{The detailed response of the decision-maker and verifier regarding the loop 1 for spatial relationship.}}
	\label{figure:spatial1}
\end{figure*}

\begin{figure*}[ht]
	\centering
	\includegraphics[width=1\textwidth]{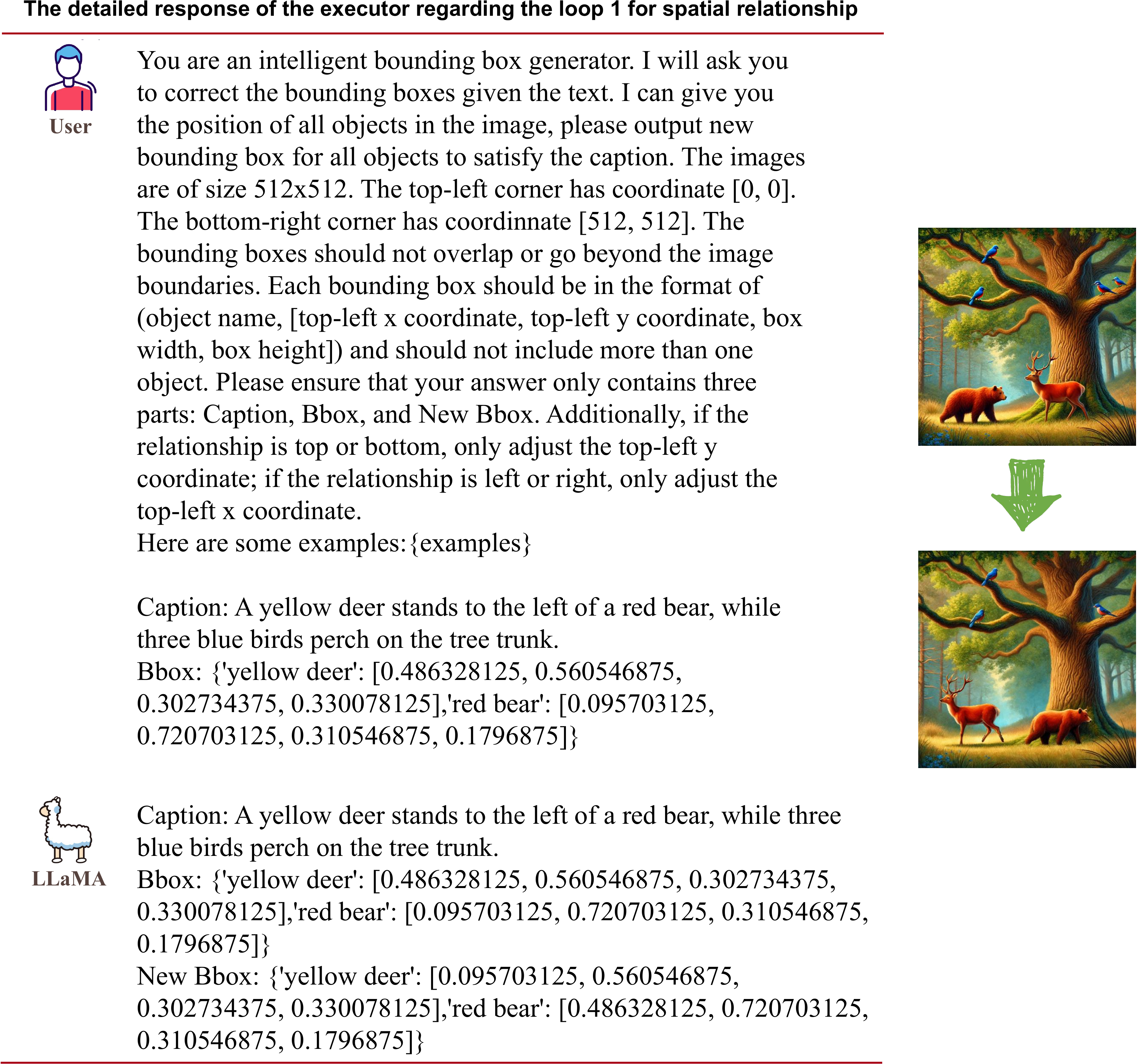} %
	\caption{\textbf{The detailed response of the executor regarding the loop 1 for spatial relationship.}}
	\label{figure:spatial2}
\end{figure*}

To enhance the interpretability and transparency of our proposed framework, we meticulously document the qualitative outcomes at each intermediate stage within the decision-execution-verification loop. As illustrated from Figures Figure~\ref{figure:deer1} to Figure~\ref{figure:spatial2}, we provide a comprehensive overview of the responses and prompts for the two-stage visual reasoning paradigm. Notably, as depicted in Figure~\ref{figure:deer1}, logical errors frequently emerge in the responses generated by LLM. This observation underscores the critical necessity of our verification mechanism in ensuring the accuracy and reliability of the overall process.

\end{document}